%% file: example_paper.tex
\documentclass{article}

\usepackage{microtype}
\usepackage{graphicx}
\usepackage{subcaption}
\usepackage{booktabs} 

\usepackage{hyperref}

\usepackage[accepted]{icml2026}

\usepackage{amsmath}
\usepackage{amssymb}
\usepackage{mathtools}
\usepackage{amsthm}

\usepackage[capitalize,noabbrev]{cleveref}

\theoremstyle{plain}
\newtheorem{theorem}{Theorem}[section]
\newtheorem{proposition}[theorem]{Proposition}
\newtheorem{lemma}[theorem]{Lemma}
\newtheorem{corollary}[theorem]{Corollary}
\theoremstyle{definition}
\newtheorem{definition}[theorem]{Definition}
\newtheorem{assumption}[theorem]{Assumption}
\theoremstyle{remark}
\newtheorem{remark}[theorem]{Remark}

\usepackage[textsize=tiny]{todonotes}

\input{header}

\icmltitlerunning{PhysHanDI: Physics-Based Reconstruction of Hand-Deformable Object Interactions}

\begin{document}

\twocolumn[
  \icmltitle{PhysHanDI: Physics-Based Reconstruction of \\
    Hand-Deformable Object Interactions}



  \icmlsetsymbol{equal}{*}

  \begin{icmlauthorlist}
    \icmlauthor{Jihyun Lee}{equal,kaist}
    \icmlauthor{Changmin Lee}{equal,kaist}
    \icmlauthor{Donghwan Kim}{kaist}
    \icmlauthor{Tae-Kyun Kim}{kaist}
  \end{icmlauthorlist}

  \icmlaffiliation{kaist}{School of Computing, KAIST, Daejeon, South Korea}

  \icmlcorrespondingauthor{Tae-Kyun Kim}{kimtaekyun@kaist.ac.kr}

  \icmlkeywords{Machine Learning, ICML}

  \vskip 0.3in
]



\printAffiliationsAndNotice{\icmlEqualContribution}

\input{sec/01_abstract}

\input{sec/02_intro}

\input{sec/03_related_work}
\input{sec/04_method}
\input{sec/05_experiments}

\input{sec/06_conclusion}

\section*{Impact Statement}
This paper presents a physics-based framework for reconstructing hand–deformable object interactions from sparse-view RGB-D data, with the goal of advancing 3D perception and physical reasoning in machine learning. The proposed method may enable downstream applications in areas such as embodied AI, digital human modeling, and immersive AR/VR content creation. Compared to prior work in RGB-D capture and human–object interaction modeling, our approach does not introduce new data modalities or privacy risks beyond those already present in existing vision-based reconstruction systems.

\section*{Acknowledgement}
This work was supported by
NST grant (CRC 21015, MSIT),
IITP grant (RS-2023-00228996, RS-2024-00459749, RS-2025-25443318, RS-2025-25441313, RS-2026-25526850, RS-2026-25522885, MSIT),
KOCCA grant (RS-2024-00442308, MCST) and
InnoCORE program (N10260110, MSIT).

\bibliography{example_paper}
\bibliographystyle{icml2026}

\newpage
\appendix
\onecolumn
\input{sec/07_appendix}

\end{document}

%% file: header.tex
\usepackage{times}

\input{math_commands.tex}

\usepackage{hyperref}
\usepackage{url}

\usepackage[utf8]{inputenc} 
\usepackage[T1]{fontenc}    
\usepackage{hyperref}       
\usepackage{url}            
\usepackage{booktabs}       
\usepackage{amsfonts}       
\usepackage{nicefrac}       
\usepackage{microtype}      
\usepackage{xcolor}         
\usepackage{comment}
\usepackage{graphicx}
\usepackage{enumitem}
\usepackage{tabularx}
\usepackage{multirow}
\usepackage{makecell}
\usepackage{marvosym}
\usepackage{adjustbox}
\newcolumntype{Y}{>{\centering\arraybackslash}X}

\usepackage{scrextend}
\deffootnote[1em]{1em}{1em}{\textsuperscript{\thefootnotemark}}
\newcommand{\jl}[1]{\noindent {\color{red}Jihyun: \textit{#1}}}
\newcommand{\cl}[1]{\noindent {\color{blue}{#1}}}

\newcommand{\keyword}[1]{{\noindent\textbf{#1}}}
\usepackage{subcaption}

%% file: math_commands.tex
\usepackage{amsmath,amsfonts,bm}

\def\eqref#1{equation~\ref{#1}}

\def\1{\bm{1}}

\DeclareMathAlphabet{\mathsfit}{\encodingdefault}{\sfdefault}{m}{sl}
\SetMathAlphabet{\mathsfit}{bold}{\encodingdefault}{\sfdefault}{bx}{n}



%% file: sec/01_abstract.tex
\vspace{-1.2\baselineskip}
\begin{abstract}  
While existing methods for reconstructing hand–object interactions have made impressive progress, they either focus on rigid or part-wise rigid objects—limiting their ability to model real-world objects (e.g., cloth, stuffed animals) that exhibit highly non-rigid deformations—or model deformable objects without full 3D hand reconstruction.
To bridge this gap, we present \textsc{PhysHanDI} (\textbf{\underline{Phys}}ics-based Reconstruction of \textbf{\underline{Hand}} and \textbf{\underline{D}}eformable Object \textbf{\underline{I}}nteractions), a framework that enables full 3D reconstruction of both interacting hands and non-rigid objects.
Our key idea is to \emph{physically simulate} object deformations driven by forces induced from densely reconstructed 3D hand motions, ensuring that the reconstructed object dynamics are both physically plausible and coherent with the interacting hand movements.
Furthermore, we demonstrate that such simulation of object deformations can, in turn, refine and improve hand reconstruction via inverse physics. In experiments, \textsc{PhysHanDI} outperforms the state-of-the-art baseline across reconstruction and future prediction.
\end{abstract}
\vspace{-2.2\baselineskip}

%% file: sec/02_intro.tex
\section{Introduction}
\label{sec:intro}

\vspace{-0.2\baselineskip}
The hand is our primary tool for interacting with objects, enabling a wide range of everyday object manipulation tasks (e.g., picking up a cell phone, folding clothes).
Effective modeling of such hand–object interactions in 3D is crucial for enabling machines to perceive and reason about human actions, which in turn is important for applications such as immersive AR/VR experiences, robot learning from human demonstrations, and teleoperation.
Owing to this importance, numerous studies have investigated the modeling hand–object interactions and reconstructing them from diverse sensing modalities, such as RGB images, depth maps, and RGB-D data~\citep{hampali2020honnotate, chao2021dexycb, hasson2019learning, mueller2017real, garcia2018first, brahmbhatt2020contactpose, taheri2020grab, fan2023arctic, swamy2023showme, corona2020ganhand, damen2022rescaling, brahmbhatt2019contactdb, garcia2020physics, antotsiou2021adversarial, kim2024mhcdiff}.

\vspace{-0.2\baselineskip}
While these existing approaches have shown impressive progress, most of them are limited to modeling interactions with \emph{rigid objects}.
Although many real-world objects (e.g., cloth, charger cables) exhibit highly non-rigid deformation, most existing methods consider rigid or part-wise rigid objects in interaction~\citep{hampali2020honnotate, chao2021dexycb, brahmbhatt2020contactpose, taheri2020grab, fan2023arctic, swamy2023showme, corona2020ganhand, damen2022rescaling, brahmbhatt2019contactdb, lee2024interhandgen, lee2023im2hands, kim2024bitt, cho2024dense}.
%
Modeling and reconstructing such object dynamics is indeed straightforward, as they can be represented by a small set of rigid transformations corresponding to each rigid body.

\vspace{-0.2\baselineskip}
In contrast, non-rigid deformation involves complex, spatially varying dynamics with substantially higher degrees of freedom, making it harder to learn reliable dynamics from input data.
While a few works have tackled hand–deformable object interaction modeling~\citep{xie2023hmdo,qi2025human,jiang2025phystwin}, most of them~\citep{xie2023hmdo,qi2025human} are limited to only small, localized deformations from \emph{finger pressure} and do not readily extend to more general, large-scale non-rigid deformations.
The most relevant work is PhysTwin~\citep{jiang2025phystwin}, which is capable of modeling large non-rigid deformations through \emph{physical simulation}. Its focus, however, is primarily on reconstructing deformable objects \emph{without full 3D hand reconstruction}. Instead, controllers are represented by only a sparse set of points (whose cardinality is about 30) directly sampled from depth maps, which may limit the precision of interaction force modeling and lead to suboptimal model topology reconstruction for simulation, as discussed later.

\vspace{-0.2\baselineskip}
To address this, we introduce \textsc{PhysHanDI} (\textbf{\underline{Phys}}ics-based Reconstruction of \textbf{\underline{Hand}} and \textbf{\underline{D}}eformable Object \textbf{\underline{I}}nteractions), a framework that enables \emph{dense 3D reconstruction} of interacting hands and non-rigid objects through physics-based simulation.
We represent hands using a dense parametric model (MANO model~\citep{romero2017embodied}) and objects using a classical physics-based model (Spring–Mass model~\citep{liu2013fast, jiang2025phystwin}) capable of simulating the dynamics of deformable objects.
In particular, the simulation of object deformation is driven by forces induced from dense motions of MANO hand meshes, enabling the modeling of object dynamics that is both physically plausible and also coherent with the fully reconstructed interacting hand movements.
%


\vspace{-0.3\baselineskip}
We propose an optimization pipeline to reconstruct this 3D dense hand–deformable object interaction model from sparse-view RGB-D videos. The pipeline consists of three stages: (1) hand reconstruction, (2) object reconstruction, and (3) hand refinement.
In the \emph{hand reconstruction} stage, we fit the MANO model~\citep{romero2017embodied} to the input RGB-D observations. 
In the \emph{object reconstruction} stage, we fit the parameters of the Spring-Mass model~\citep{liu2013fast, jiang2025phystwin} conditioned on the reconstructed 3D hands.
In particular, we simulate object deformations via spring–mass system~\citep{liu2013fast, jiang2025phystwin} driven by interaction forces induced from the reconstructed hand motions, and the resulting simulated object geometry is compared against the input RGB-D observations for parameter optimization.
%
%
In the final \emph{hand refinement} stage, we refine the initial hand reconstructions via inverse physics, leveraging the physics-based object model fitted in the previous stage. This refinement enforces that the reconstructed hands produce object simulations that are more consistent with the input observations. While we empirically find that the initial hand reconstruction stage is already sufficient to achieve state-of-the-art results with multi-view RGB-D inputs, this hand refinement stage proves especially effective when inference is performed from sparser inputs (e.g., future prediction in a single-view setting). To the best of our knowledge, this is the \textit{\textbf{first}} work to demonstrate that inverse physics, guided by a physics-based deformable object model, can enhance hand reconstruction.

%
\vspace{-0.3\baselineskip}
To experimentally validate the effectiveness of our method, we compare it against state-of-the-art baselines~\citep{jiang2025phystwin,zhang2024dynamics,zhong2024reconstruction} for physics-based reconstruction of deformable objects, and demonstrate that our method outperforms them in reconstruction and future prediction.

%
%

\vspace{-0.3\baselineskip}
Our contributions can be summarized as follows:
\vspace{-0.4\baselineskip}
\begin{itemize}
  \vspace{-0.5\baselineskip}
  \item We present \textsc{PhysHanDI}, a framework for reconstructing hand–deformable object interactions through physical simulation. To the best of our knowledge, \textsc{PhysHanDI} is the first approach to achieve dense 3D reconstruction of both hands and deformable objects from sparse-view RGB-D videos.
  \vspace{-0.5\baselineskip}
  \item For deformable objet reconstruction, we propose to simulate object deformations driven by interaction forces induced from fully reconstructed 3D hand motions to achieve more accurate simulation than the existing state-of-the-art~\citep{jiang2025phystwin} based on a sparse hand representation.
  \vspace{-0.3\baselineskip}
  \item For hand reconstruction, we refine the initial MANO~\citep{romero2017embodied} fitting through inverse physics, leveraging the previously reconstructed physics-based object model. To the best of our knowledge, this is the first work to show that inverse physics, guided by a physics-based deformable object model, can improve hand reconstruction.
  \vspace{-0.3\baselineskip}
  \item We achieve new state-of-the-art performance compared to our most relevant approach, PhysTwin~\citep{jiang2025phystwin}, in reconstruction and future prediction.
\end{itemize}

%% file: sec/03_related_work.tex
\begin{figure*}[t]
\begin{center}
\includegraphics[width=0.85\textwidth]{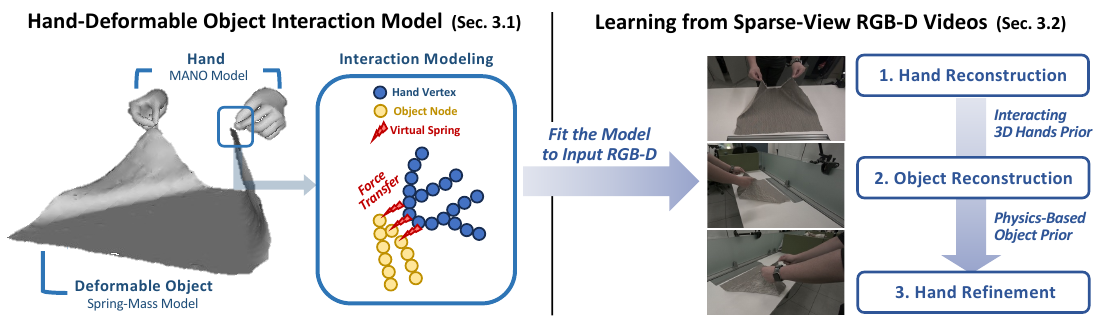}
\caption{\textbf{\textsc{PhysHanDI} models physically plausible hand–deformable object interactions.} In our interaction model, each hand is represented by the MANO model~\citep{romero2017embodied}, and each object is represented by a spring–mass model~\citep{liu2013fast}. Their interaction is modeled by simulating object deformations driven by interaction forces derived from the reconstructed 3D hand motions. Our interaction model can be learned from sparse-view RGB-D videos through three stages: (1) hand reconstruction, (2) object reconstruction, and (3) hand refinement.} 
\label{fig:method_overview}
\end{center}
\vspace{-1.2\baselineskip}
\end{figure*}

\vspace{-\baselineskip}
\section{Related Work}
\label{sec:related_work}

\subsection{3D Hand-Object Interaction Modeling}
\keyword{Hand and \textit{rigid} object interaction.}
There are numerous works on modeling and reconstructing hand–object interactions from various types of inputs, e.g., RGB, depth, or RGB-D~\citep{chen2021joint,liu2021semi,doosti2020hope,hasson2020leveraging,hasson2019learning,hampali2022keypoint,tekin2019h,chen2022alignsdf,chen2023gsdf}, or on estimating hand-object contacts to support such reconstruction~\citep{tse2022s,jung2025learning}. Most of these methods assume a rigid object in interaction, where the object dynamics is represented with a reference shape (e.g., a given template shape or a reconstructed shape from the first frame) with global rigid transformation~\citep{chen2021joint,liu2021semi,doosti2020hope,hasson2020leveraging,hasson2019learning,hampali2022keypoint,tekin2019h,chen2022alignsdf,chen2023gsdf}.
Recently, there have been efforts to model part-wise rigid objects under hand interactions, where the object is additionally represented with part labels and per-part rigid transformations~\citep{fan2023arctic,zhu2024contactart,zhang2025bimart}.
While this enables more expressive deformation modeling than prior works with a global rigidity assumption, these methods remain non-trivial to extend to more general real-world objects (e.g., cloth, charger cables) that exhibit non-rigid deformations.
%
%

\keyword{Hand and \textit{non-rigid} object interaction.}
There are only a few works that attempt to model and reconstruct hand–\emph{non-rigid} object interactions.
HMDO~\citep{xie2023hmdo} proposes a pipeline for markerless capture of hand–deformable object interactions from multi-view images.
However, its main focus is on modeling \emph{localized deformations driven by finger pressure}, and “\emph{the interacting objects in [its] dataset do not have large deformations, such as 180-degree twisting or bending}”~\citep{xie2023hmdo}. Therefore, it is non-trivial to apply this method to our targeted hand–object interaction datasets, where large, global non-rigid deformations occur (e.g., bending a doll’s arms).
Similarly, a recent work on generating hand–deformable object interactions~\citep{qi2025human} also assumes that object deformations are locally driven by finger pressure, based on the HMDO dataset.
A concurrent line of work on hand–object contact estimation for nonrigid objects~\citep{xie2023nonrigid} similarly targets localized contact regions rather than modeling global deformation dynamics.

The most related work to ours is PhysTwin~\citep{jiang2025phystwin}, a recent state-of-the-art method for physics-based deformable object reconstruction from multi-view RGB-D videos that can handle non-localized deformations.
While it can address hand–deformable object interaction scenarios, its main focus is on object modeling, with the interactee represented only by sparse points directly fetched from input depth maps; in this case, true hand–object contact points are unobservable due to contact occlusions. In Sec.~\ref{sec:experiments}, we further demonstrate that our method produces more accurate object reconstruction, and that the reconstructed object model can, in turn, refine the initial hand reconstruction—mutually benefiting each other.

\vspace{-0.3\baselineskip}

\subsection{Deformable Object Modeling}

\vspace{-0.3\baselineskip}
\keyword{Dynamic reconstruction-based modeling.}
Dynamic reconstruction-based methods recover 3D representations (e.g., Occupancy Functions~\citep{mescheder2019occupancy}, Neural Radiance Fields~\citep{mildenhall2020nerf}, 3D Gaussian Splats~\citep{park20203d}) from inputs such as RGB~\citep{attal2023hyperreel, kratimenos2024dynmf, li2023dynibar, luiten2024dynamic, park2021nerfies, park2021hypernerf, pumarola2021d, wang2023flow, xian2021space, yu2023dylin, tretschk2021nonrigid, chu2022physics}, depth~\citep{curless1996volumetric, li2008global}, or RGB-D~\citep{newcombe2015dynamicfusion} data.
Most recent methods typically reconstruct a canonical representation (e.g., at the first frame) and learn deformation fields to capture object dynamics~\citep{park2021nerfies,park2021hypernerf,kratimenos2024dynmf,xian2021space}.
Despite differences in exact modeling approaches, they share a key limitation: the focus remains on \emph{reconstructing} 3D representations that match observed inputs, without explicitly modeling physical properties—thereby limiting their ability to support future prediction or generalization to unseen predictions, as also discussed in~\citep{jiang2025phystwin}.

\vspace{-0.1\baselineskip}
\keyword{Simulation-based modeling.}
Simulation-based methods enable the modeling of object dynamics in a physically plausible manner, while also allowing generalization to unseen interactions. %
Early works relied on pre-scanned static objects and clean point clouds~\citep{wang2015deformation, Qiao2021Differentiable, du2021diffpd, geilinger2020add, jatavallabhula2021gradsim}, or were constrained to synthetic data or highly dense viewpoints~\citep{zhang2024physdreamer, li2023pac, chen2022virtual, zhong2024reconstruction, qiao2022neuphysics}.
%
More recent methods~\citep{zhang2024dynamics, jiang2025phystwin, yang2025physworld, xu2026neuspring} take sparse-view real RGB-D images as input, thereby reducing the burden of expensive capture setups. However, none of these methods explicitly model the full 3D geometry of the interactee; instead, hand–object interactions are represented through sparse control signals, which can adversely affect the fidelity of deformable object reconstruction under complex contact scenarios.

%
%
%
%

\vspace{-0.5\baselineskip}

%% file: sec/04_method.tex

\section{\textsc{PhysHanDI}}
\label{sec:method}

In this section, we first introduce our physics-based model for dense hand–deformable object interaction (Sec.~\ref{subsec:representation}).
We then describe how this model can be reconstructed from sparse-view RGB-D videos (Sec.~\ref{subsec:handi_training}).
%
\vspace{-0.2\baselineskip}

\subsection{Physics-Based Interaction Modeling}
\label{subsec:representation}
\vspace{-0.2\baselineskip}
We now present our approach to physically based modeling of hand–deformable object interactions.
%
We first describe how the hand and object are \emph{each} represented, and then elaborate on how their interaction is modeled through physics-based simulation.

\keyword{Hand Representation: MANO Model~\citep{romero2017embodied}.}
We represent each hand by the parameters of MANO model~\citep{romero2017embodied}, a widely used PCA-based hand model.
It maps a pose parameter $\boldsymbol{\theta} \in \mathbb{R}^{45}$, a shape parameter $\boldsymbol{\beta} \in \mathbb{R}^{10}$, a global rotation $\mathbf{R} \in SO(3)$ and a translation $\mathbf{t} \in \mathbb{R}^3$ to a dense 3D hand mesh $\mathcal{M} = (\mathcal{V}, \mathcal{F})$ with vertices $\mathcal{V} = \{\mathbf{v}_{i}\}_{i=1}^{778}$ and triangular faces $\mathcal{F} = \{\mathbf{f}_{i}\}_{i=1}^{1554}$.
Since the model provides a prior that constrains the solution space of 3D hand meshes within the low-dimensional parameter space, it has been widely adopted to reduce ill-posedness in various hand reconstruction problems (e.g., interacting hand and rigid object reconstruction).

\keyword{Object Representation: Spring-Mass Model~\citep{liu2013fast, jiang2025phystwin}.}
We represent each deformable object using a spring-mass model~\citep{liu2013fast, jiang2025phystwin}, a classical physics-based model capable of simulating the dynamic behavior of deformable objects. It models an object as a graph $\mathcal{O} = (\mathcal{N}, \mathcal{E})$. $\mathcal{N} = \{\mathbf{n}_{i}\}_{i=1}^{N}$ denotes a set of $N$ number of mass nodes, where each mass node $\mathbf{n}_{i}$ is parameterized by its position $\mathbf{x}_{i} \in \mathbb{R}^{3}$ and velocity $\mathbf{v}_{i} \in \mathbb{R}^{3}$\footnote{While the hand mesh vertex is also denoted by $\mathbf{v}_{i}$, we allow a slight abuse of notation to remain consistent with notation conventions used in related work.}, and mass $m_{i} \in \mathbb{R}$\footnote{Directly following prior work~\citep{jiang2025phystwin}, we assign a unit mass to all nodes in the spring–mass system, since ground-truth mass values are not available in our setting.}.
$\mathcal{E} = \{(i, j)\: |\: i, j \in \{1, ..., N \}\}$ denotes a set of springs connecting the mass nodes, where $i$ and $j$ are the indices of the mass nodes connected by each spring.
In this spring-mass model, each mass node can be simulated in response to the force acting on it. In particular, the force on each mass node $\mathbf{n}_{i}$ is modeled as:

\vspace{-\baselineskip}
\begin{equation}
\mathbf{F}_i = \sum_{(i,j) \in \mathcal{E}} \mathbf{F}_{i,j}^{\text{spring}} + \mathbf{F}_{i,j}^{\text{damping}} + \mathbf{F}_i^{\text{external}}.
\label{eq:spring_mass_force}
\end{equation}
\vspace{-\baselineskip}

\noindent The first term $\mathbf{F}_{i,j}^{\text{spring}} = s_{ij} \left( \|\mathbf{x}_j - \mathbf{x}_i\| - r_{ij}\right) 
\frac{\mathbf{x}_j - \mathbf{x}_i}{\|\mathbf{x}_j - \mathbf{x}_i\|}$ represents the spring force between the connected mass nodes $\mathbf{n}_i$ and $\mathbf{n}_j$ based on Hooke’s law, where $s_{ij}$ is the stiffness parameter, and $r_{ij}$ is the rest length of the spring $(i, j)$.
This term encourages the spring-mass system to maintain the rest length of each spring.
The second term $\mathbf{F}_{i,j}^{\text{damping}} = -\gamma_{ij} (\mathbf{v}_i - \mathbf{v}_j)$ is a dashpot damping force between $\mathbf{n}_{i}$ and $\mathbf{n}_{j}$, where $\gamma_{ij}$ is the dashpot damping coefficient of the spring $(i, j)$. 
It penalizes relative velocity along the spring direction, stabilizing the system and preventing oscillations. The final term $\mathbf{F}_i^{\text{external}}$ models external forces acting on the mass node, such as gravity. 

Given the force $\mathbf{F}_{i}$ computed from the above modeling equation (Eq.~\ref{eq:spring_mass_force}) at each time $t$, the updated position $\mathbf{x}_{i}$ of node $i$ at time $t+1$ is obtained by numerically integrating Newton’s second law over time, such that 
$\mathbf{v}_i^{t+1} = \mathbf{v}_i^t + \Delta t \frac{\mathbf{F}_i}{m_i}$ and  
$\mathbf{x}_i^{t+1} = \mathbf{x}_i^t + \Delta t \, \mathbf{v}_i^{t+1}$.


\keyword{Hand-Deformable Object Interaction Modeling.}
Given these hand and deformable object representations, we now describe how their interaction is modeled by simulating object deformation with a spring–mass system, driven by forces induced by the motions of MANO hand meshes.
We follow the common strategy for modeling interaction forces in the spring–mass system~\citep{liu2013fast}, where \emph{virtual springs} are connected between object nodes and the interactee (MANO hand vertices in our case) that are detected to be in contact within a connection radius $\delta$ (left subfigure of Fig.~\ref{fig:method_overview}).

Formally, in our final spring--mass system, the mass nodes are defined as $\mathcal{N} \cup \mathcal{V}'$, which is the superset of the object nodes $\mathcal{N}$ and the \emph{virtual hand nodes} $\mathcal{V}'$. Since these virtual hand nodes are used to induce forces for simulating object deformation, their positions and velocities are determined by the tracked MANO vertices $\mathcal{V}$, and then fixed as a boundary condition throughout the simulation. The virtual springs are then defined as $\mathcal{E} \cup \mathcal{E}^{\text{virtual}}$, which is the superset of the object springs $\mathcal{E}$ and the virtual springs $\mathcal{E}^{\text{virtual}}$ connecting the contacted object and hand nodes.
Note that, during simulation, these virtual springs encourage the object regions 
in contact with the hand to smoothly deform according to the fixed hand vertex motion, as 
the spring and dashpot damping forces in the spring--mass system 
($\mathbf{F}_{i,j}^{\text{spring}}$ and $\mathbf{F}_{i,j}^{\text{damping}}$ in 
Eq.~\ref{eq:spring_mass_force}) act to maintain the contact topology 
(i.e., the rest length of the \emph{virtual springs} between hand vertices and 
object nodes). This ensures object dynamics that 
are physically plausible and coherent with the interacting hand movements, while 
being capable of modeling large and complex non-rigid deformations.

\begin{figure}[t]
\begin{center}
\includegraphics[width=\columnwidth]{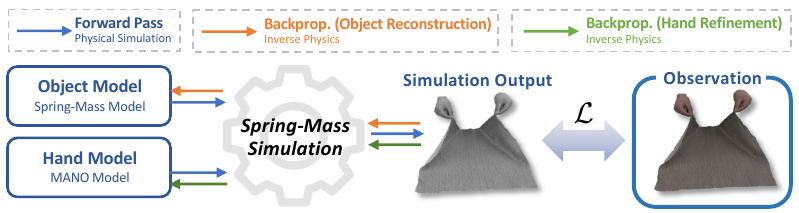}
\caption{\textbf{Illustration of inverse physics for object reconstruction and hand refinement.} Our spring–mass simulation is driven by the spring–mass object model and the MANO hand model. In the object reconstruction stage, the object model is fitted via inverse physics given the initial MANO models, while in the subsequent hand refinement stage, the initial MANO models are refined given the reconstructed object model. $\mathcal{L}$ denotes our loss function, composed of $\mathcal{L}_{\textit{ch}}$ and $\mathcal{L}_{\textit{tr}}$.} 
\label{fig:inverse_physics_overview}
\end{center}
\vspace{-2\baselineskip}
\end{figure}

\vspace{-0.8\baselineskip}
\subsection{Learning from Sparse-View RGB-D Videos}
\label{subsec:handi_training}
\vspace{-0.5\baselineskip}

We now explain how the physics-based hand–deformable object interaction model described in Sec.~\ref{subsec:representation} can be reconstructed from sparse-view RGB-D video inputs. Our learning pipeline consists of three stages: (1) hand reconstruction, (2) object reconstruction, and (3) hand refinement.

%
\vspace{-0.2\baselineskip}
\keyword{Hand Reconstruction.}
In this stage, we fit the MANO hand model~\citep{romero2017embodied} to multi-view RGB-D videos. 
Our optimization target for each frame is the MANO parameters 
$\Theta_{h} = \{ \boldsymbol{\theta}, \boldsymbol{\beta}, \mathbf{R}, \mathbf{t} \}$, 
where $\boldsymbol{\theta} \in \mathbb{R}^{45}$ and $\boldsymbol{\beta} \in \mathbb{R}^{10}$ are pose and shape parameters, and 
$\mathbf{R} \in SO(3)$ and $\mathbf{t} \in \mathbb{R}^3$ are global rotation and translation.    
Our optimization objective is formulated as:  

\vspace{-2\baselineskip}
\begin{equation}
\hspace{-5pt} \min_{\Theta_{h}} \;
\mathcal{L}_{2D}(\Theta_{h}, \mathbf{U}) 
+ \lambda_{d} \, \mathcal{L}_{d}(\Theta_{h}, \mathbf{D}) 
+ \lambda_{t} \, \mathcal{L}_{t}(\Theta_{h}, \Theta_{h}^{\textit{prev}}),
\label{eq:hand_recon_loss}
\end{equation}
\vspace{-1.5\baselineskip}

where $\mathcal{L}_{2D}$ measures the reprojection error between the projected MANO keypoints and the 2D keypoint supervision $\mathbf{U} \in \mathbb{R}^{V \times 21 \times 2}$ for each of the $V$ views. 
$\mathcal{L}_{d}$ measures the discrepancy between the rendered MANO depth and the observed depth maps $\mathbf{D} \in \mathbb{R}^{V \times H \times W}$, where $H$ and $W$ denote the depth map resolution, and $\mathcal{L}_{t}$ regularizes the temporal smoothness of the MANO parameters with respect to those fitted in the previous frame, $\Theta_{h}^{\textit{prev}}$. The coefficients $\lambda_{2D}$, $\lambda_{d}$, and $\lambda_{t}$ control the relative weight of each loss term.

\keyword{Object Reconstruction.}
Conditioned on the fitted 3D hands, we now fit the spring–mass model representing the deformable object under interaction.
For this stage, we mainly follow the object model fitting pipeline of PhysTwin~\citep{jiang2025phystwin}, though it only considers the sparse controller points as the interactee.
At a high level, the object’s 3D geometry at $t=0$ is first obtained using an image-to-3D generative model~\citep{xiang2025structured}.
The object dynamics for $t \in [1, T]$ are then simulated with a spring–mass system, and the physical parameters (e.g., $s_{ij}$, $\gamma_{ij}$ and $\delta$ in Sec.~\ref{subsec:representation}) are optimized so that the simulated geometries better match the input observations. The optimization objective is formulated as minimizing two terms: (1) $\mathcal{L}_{\textit{ch}}$, which measures the Chamfer distance between the simulated node positions and the observed 3D point clouds lifted from the input depth maps, and (2) $\mathcal{L}_{\textit{tr}}$, an $\ell_2$ loss between the simulated node positions and the pseudo–ground-truth 3D points tracked by CoTracker3~\citep{karaev2024cotracker3}.

While we kindly refer the reader to our supplementary material or \cite{jiang2025phystwin} for more details on this stage, we highlight a key difference in our spring–mass simulations: our approach models interaction forces from the dense 3D hand geometry fitted with the MANO model in the previous stage, whereas PhysTwin approximates these forces using sparse points sampled from input depth maps, where true hand–object contact points are unobservable due to contact occlusions. As this may limit the precision of interaction force modeling during simulation, our method achieves more accurate simulation results, as discussed in Sec.~\ref{sec:experiments}. In Sec.~\ref{subsec:contact_analysis}, we also provide numerical analysis showing that our dense hand interactee enables the spring–mass model topology to be reconstructed more optimally than PhysTwin in the view of peridynamics~\citep{silling2005meshfree,silling2007peridynamic,wang2023determination}.

\input{tab/main_comp_dynamic_CD_dh}
\input{tab/main_comp_our_dataset_dh}

\keyword{Hand Refinement.}
After the object reconstruction stage, we can leverage the reconstructed physics-based object model as an additional prior to further refine the initial hand model fitting, enforcing it to produce \emph{object simulations} better aligned with the input observation.
Specifically, we reuse the same $\mathcal{L}_{\textit{ch}}$ and $\mathcal{L}_{\textit{tr}}$ losses from the object reconstruction stage to measure the discrepancy between the simulated object nodes, and the ground-truth observations at each timestep $t$. In this stage, however, we apply them to fine-tune the MANO model parameters via inverse physics (see Fig.~\ref{fig:inverse_physics_overview}), using gradient-descent-based optimization.
Let $\mathcal{S}_{t}(\cdot)$ denote the function that returns the simulated object nodes at timestep $t$ given the MANO hand parameters. The refined hand parameters $\tilde{\Theta}_{h}$ are then optimized as:


\vspace{-1.8\baselineskip}
\begin{gather}
\tilde{\Theta}_{h} = \arg\min_{\Theta_{h}} \frac{1}{T} \sum_{t=1}^T \mathcal{L}(\Theta_{h}, \mathcal{P}, \mathbf{T}), \\
\hspace{-5pt} \mathcal{L}(\Theta_{h}, \mathcal{P}, \mathbf{T}) = \mathcal{L}_{\textit{ch}}(\mathcal{S}_{t}(\Theta_{h}), \mathcal{P})
+ \lambda_{\textit{tr}} 
\mathcal{L}_{\textit{tr}}(\mathcal{S}_{t}(\Theta_{h}), \mathbf{T}),
\label{eq:loss_function}
\end{gather}
\vspace{-1.8\baselineskip}


\noindent where $\mathcal{P}$ and $\mathbf{T}$ denote the ground-truth lifted point cloud and tracked points, respectively.
This inverse-physics–based refinement is particularly effective when the hand observation is highly ill-posed; while we empirically find that the initial hand reconstruction stage is already sufficient to achieve state-of-the-art results with multi-view RGB-D inputs, this hand refinement stage proves especially effective when inference is performed from sparser inputs (e.g., future prediction in a single-view setting).
To the best of our knowledge, this is the first work to demonstrate that hand model fitting accuracy can be improved through the inverse physics of deformable object simulation.

\vspace{-0.5\baselineskip}


%% file: tab/main_comp_dynamic_CD_dh.tex
\begin{table*}[!h]
\centering
\small
\resizebox{\linewidth}{!}{%
\begin{tabular}{c|ccc|cc|ccc|cc}
\toprule
\multirow{3}{*}{\textbf{Method}} & \multicolumn{5}{c|}{\textbf{Reconstruction \& Resimulation}} & \multicolumn{5}{c}{\textbf{Future Prediction}} \\
\cmidrule(lr){2-11}
\multirow{2}{*}{} & \multicolumn{3}{c|}{\textbf{3D Metrics}} & \multicolumn{2}{c|}{\textbf{2D Metrics}} & \multicolumn{3}{c|}{\textbf{3D Metrics}} & \multicolumn{2}{c}{\textbf{2D Metrics}} \\
 & $\text{CD}_\text{dyn}$ $\downarrow$ & $\text{CD}_\text{full}$ $\downarrow$ & Track Err. $\downarrow$ & IoU $\uparrow$ & PSNR $\uparrow$ & $\text{CD}_\text{dyn}$ $\downarrow$ & $\text{CD}_\text{full}$ $\downarrow$ & Track Err. $\downarrow$ & IoU $\uparrow$ & PSNR $\uparrow$ \\
\midrule
Spring-Gaus~\cite{zhong2024reconstruction} & 27.79 & 38.84 &	4.65&	0.55&	21.41&	37.38 &	56.51& 7.69&	0.42&	19.91 \\
GS-Dynamics~\cite{zhang2024dynamics} & 33.37 & 13.67& 1.81& 0.74& 23.12& 56.79 & 33.99 & 4.50& 0.51& 18.97\\
PhysTwin~\cite{jiang2025phystwin} & 10.78 & 5.90& 	1.00&	0.84&	25.23&	16.32 &	11.45& 2.10&	0.70&	22.07 \\
\textbf{\textsc{PhysHanDI} (Ours)} &	\textbf{8.32}& \textbf{5.30}& \textbf{0.89}& \
\textbf{0.85}&	\textbf{25.62}&	\textbf{14.35}&	\textbf{10.57}&	\textbf{2.05}&	\textbf{0.73}&	\textbf{22.84}\\

\bottomrule
\end{tabular}
}
\vspace{0.5\baselineskip}
\caption{\textbf{Reconstruction \& Resimulation and Future Prediction results on the PhysTwin-dense dataset~\citep{jiang2025phystwin}}. Our method outperforms the state-of-the-art~\citep{jiang2025phystwin} on all metrics. $\textrm{CD}$ is measured in millimeters, and $\textrm{Track Err.}$ is scaled by $\times100$ for readability.} 
\label{tab:recon_future_phystwin_dataset}
\vspace{-1.5\baselineskip}
\end{table*}

%% file: tab/main_comp_our_dataset_dh.tex
\begin{table*}[!h]
\centering
\fontsize{5pt}{5pt}\selectfont
\resizebox{\linewidth}{!}{%
\begin{tabular}{c|cc|cc|cc|cc}
\toprule
\multirow{3}{*}{\textbf{Method}} & \multicolumn{4}{c|}{\textbf{Reconstruction \& Resimulation}} & \multicolumn{4}{c}{\textbf{Future Prediction}} \\
\cmidrule(lr){2-9}
\multirow{2}{*}{} & \multicolumn{2}{c|}{\textbf{3D Metrics}} & \multicolumn{2}{c|}{\textbf{2D Metrics}} & \multicolumn{2}{c|}{\textbf{3D Metrics}} & \multicolumn{2}{c}{\textbf{2D Metrics}} \\
 & CD $\downarrow$ & Track Err. $\downarrow$ & IoU $\uparrow$ & PSNR $\uparrow$ & CD $\downarrow$ & Track Err. $\downarrow$ & IoU $\uparrow$ & PSNR $\uparrow$ \\

\midrule
PhysTwin~\cite{jiang2025phystwin} & 5.59 &	1.58&	0.76&	21.22&	7.98 &	2.42&	0.63&	\textbf{19.76} \\
\textbf{\textsc{PhysHanDI} (Ours)} & \textbf{5.06}&	\textbf{1.50}& \textbf{0.78}&	\textbf{21.61}&	\textbf{7.54}&	\textbf{2.40}&	\textbf{0.65}&	19.75\\
\bottomrule
\end{tabular}
}
\vspace{0.5\baselineskip}

\caption{\textbf{Reconstruction \& Resimulation and Future Prediction results on the \textsc{DenseHDI} dataset}. Our method outperforms the state-of-the-art~\citep{jiang2025phystwin} on most metrics, demonstrating its effectiveness. $\textrm{CD}$ is measured in millimeters, and $\textrm{Track Err.}$ is scaled by $\times100$ for readability.}
\label{tab:recon_future_our_dataset}
\vspace{-2.5\baselineskip}
\end{table*}

%% file: sec/05_experiments.tex
\section{Experiments}
\label{sec:experiments}

In this section, we experimentally evaluate the effectiveness of our method.
We first describe our experimental settings in Sec.~\ref{subsec:experiment_settings}, and then present the comparison results in Sec.~\ref{subsec:experimental_comparisons}. We additionally evaluate robustness under noisy input signals in Sec.~\ref{subsec:robustness_analysis}, and provide a comparative analysis of the spring--mass model topology between our method and PhysTwin~\citep{jiang2025phystwin} in Sec.~\ref{subsec:contact_analysis}.

\vspace{-0.5\baselineskip}
\subsection{Experiment Settings}
\label{subsec:experiment_settings}


\keyword{Datasets.}
We use the PhysTwin dataset~\citep{jiang2025phystwin}, which consists of three-view RGB-D videos of hand–deformable object interactions. In the original PhysTwin dataset, a non-negligible portion of the sequences contains only very sparse \emph{point-based} contacts between the hand and the object (e.g., fingers pinching the object), which are less common in practical scenarios. Since we are interested in modeling more realistic hand–object interactions, we primarily perform evaluation on a subset of the PhysTwin dataset that excludes such sequences with only point-based contacts, which we term the \emph{PhysTwin-dense} dataset -- while also presenting full results in the supplementary material.

In addition, we newly collect 19 sequences specifically designed to capture denser hand–object contacts, which we refer to as the \textsc{DenseHDI} dataset. This dataset is collected using the same data acquisition protocol as~\cite{jiang2025phystwin} and includes 10 additional objects (e.g., pouch, towel, paper cup, and hat; see the supplementary material for details). Upon publication, we will release this dataset to facilitate future research on modeling hand–deformable object interactions.

%
%
%

%

\keyword{Tasks and Baselines.}
We follow the evaluation protocol used in PhysTwin~\citep{jiang2025phystwin} and consider two evaluation tasks: (1) reconstruction and resimulation and (2) future prediction. We also report results on generalization to unseen interactions in the supplementary material.
For baselines, we compare against PhysTwin~\citep{jiang2025phystwin}, Spring-Gaus~\citep{zhong2024reconstruction}, and GS-Dynamics~\citep{zhang2024dynamics}, which are the current state of the arts in physics-based object reconstruction. Note that for Spring-Gaus, we use the controller-augmented variant adopted in the PhysTwin's comparisons, as its original formulation does not support external control inputs.
%

\vspace{-0.2\baselineskip}
\keyword{Evaluation metrics.}
We again follow the evaluation metrics used in PhysTwin~\citep{jiang2025phystwin} to ensure fair comparisons. In particular, we use metrics that evaluate geometric and photometric discrepancies between the reconstructed objects and the ground truth—either in 3D space (Chamfer Distance, Tracking Error) or in projected 2D space (IoU, PSNR). 
To enable the use of photometric metrics (PSNR), we follow PhysTwin by learning surrogate Gaussian splats bounded to the object model, which allows rendering of the reconstructed object dynamics for evaluation.
We also note that the optimization-based reconstruction pipelines in both PhysTwin~\citep{jiang2025phystwin} and our method (Sec.~\ref{subsec:handi_training}) involve stochasticity (e.g., random parameter initialization). Therefore, we run the official implementation of PhysTwin and compare the average results over 10 runs for more reliable comparisons.

\vspace{-0.5\baselineskip}

\subsection{Experimental Comparisons}
\label{subsec:experimental_comparisons}

\begin{figure*}[!t]
\begin{center}
\includegraphics[width=0.88\textwidth]{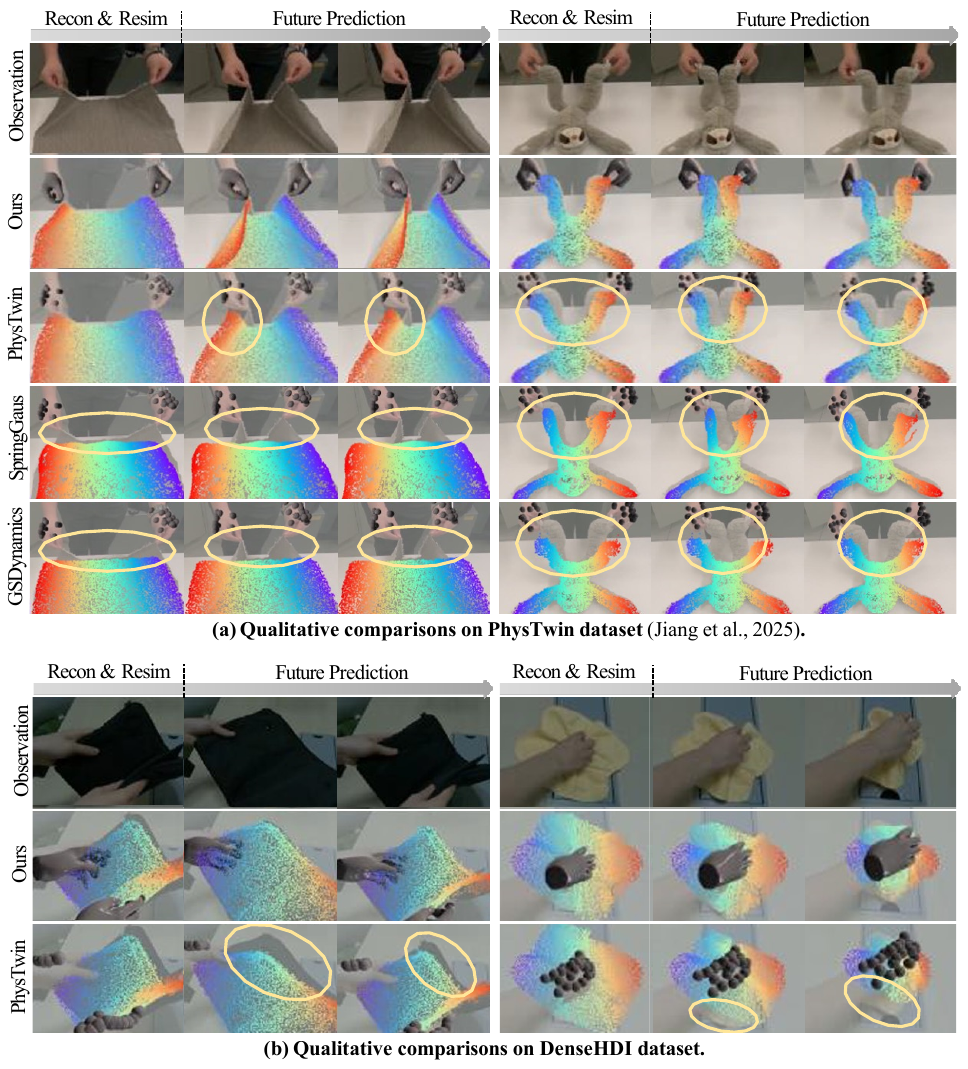}
\vspace{-0.4\baselineskip}
\caption{\textbf{Qualitative comparisons on (1) reconstruction and resimulation, and (2) future prediction.} Yellow circles indicate regions where object simulations are less accurately aligned with the ground-truth observations or with the interacting hand contacts. Compared to all the baselines, our method produces more accurate object simulations. More qualitative results are provided in the supplementary video.} 
\label{fig:handi_results}
\end{center}
\vspace{-1.5\baselineskip}
\end{figure*}

\vspace{-0.2\baselineskip}
\subsubsection{Reconstruction \& Resimulation}
\label{subsec:recon_resim}

\vspace{-0.2\baselineskip}
In the reconstruction and resimulation experiments, we evaluate reconstruction accuracy on the \emph{seen} frames used during physics-based object model fitting, following the protocol in \cite{jiang2025phystwin}.
%
%
Tab.~\ref{tab:recon_future_phystwin_dataset} (left) presents our quantitative results on the PhysTwin-dense dataset.
For this dataset, we observe that many sequences contain large static regions, with only small local regions undergoing meaningful deformation (e.g., the cloth region fixed on the table in Fig.~\ref{fig:handi_results}(a)). To more clearly measure accuracy in the dynamically deforming regions—which are the primary focus of physics-based reconstruction—we report two versions of the Chamfer Distance; $\text{CD}_\text{full}$, which is the standard Chamfer Distance computed over all object points, whereas $\text{CD}_\text{dyn}$ evaluates only the points that exhibit non-negligible deformation. Formally, given pseudo-ground-truth point trajectories $\mathbf{x}_{1:T}$ obtained from CoTracker and depth input, we assign a point in the deforming set if $\lVert \mathbf{x}_i - \mathbf{x}_1 \rVert^2 > \tau_\text{dyn}$, where $\tau_\text{dyn}$ is a motion-magnitude threshold. 
%
In the table, our method outperforms Spring-Gaus and GS-Dynamics by a large margin across all metrics, and also outperforms PhysTwin on all metrics. The qualitative comparisons in Fig.~\ref{fig:handi_results} and the supplementary video further demonstrate that our method achieves more accurate reconstruction and resimulation.
Note that Spring-Gaus was originally proposed for input settings with denser viewpoints; under our three-view sparse-input configuration, its simulation becomes unstable, leading to broken object geometry, as shown in the qualitative results.
GS-Dynamics is designed to leverage long motion sequences through its GNN-based motion representation; consequently, it fails to capture meaningful deformation behavior in shorter sequences and instead models only subtle motions.
For PhysTwin, interaction forces during simulation are approximated from sparse points (with a cardinality of around 30) sampled from depth maps. This can lead to less accurate simulations because (1) the precision of interaction force modeling is limited, as true hand–object contact points cannot be fully observed from depth sensors due to mutual occlusion, and (2) the reconstructed model topology used for simulation is suboptimal, as analyzed in Sec.~\ref{subsec:contact_analysis}.
%
%
%

%
%

In Tab.~\ref{tab:recon_future_our_dataset} (left), we additionally report results on our \textsc{DenseHDI} dataset, which mainly captures dense hand–object interactions. Here, our method again outperforms the baseline on all metrics, further validating its effectiveness in modeling dense hand–deformable object interactions through full 3D hand modeling.

\vspace{-0.5\baselineskip}
\subsubsection{Future Prediction}
\label{subsec:future_pred}

In the future prediction experiments, we evaluate reconstruction quality on future frames that were \emph{unseen} during physics-based object model fitting.

\keyword{Three-View RGB-D Inputs.}
In Tables~\ref{tab:recon_future_phystwin_dataset} and \ref{tab:recon_future_our_dataset} (right), we present future prediction results on the PhysTwin and \textsc{DenseHDI} datasets, respectively.
Our method outperforms Spring-Gaus~\citep{zhong2024reconstruction} and GS-Dynamics~\citep{zhang2024dynamics} by a substantial margin across all metrics, and outperforms PhysTwins~\citep{jiang2025phystwin} on most metrics, demonstrating its effectiveness for future prediction as well.
Our qualitative comparisons in Fig.~\ref{fig:handi_results} and our supplementary video also show that our approach produces future predictions that are more accurately aligned with the ground-truth observations.



\vspace{-0.2\baselineskip}
\keyword{Single-View RGB-D Inputs.}
We additionally report future prediction results on \emph{single-view RGB-D inputs}, which represent a more challenging scenario than the multi-view setting considered in the existing state-of-the-art (PhysTwin~\citep{jiang2025phystwin}).
In this setting, PhysTwin must approximate interaction forces from sparse hand points sampled from a \emph{single-view} depth map, which are highly partial. We empirically observed that its object simulation frequently fails due to errors in identifying hand–object contact points (determined by the threshold $\delta$ in Sec.~\ref{subsec:representation}), and therefore could not be included in our comparisons.
As shown in Tab.~\ref{tab:single_view_future_pred}, our method robustly addresses the challenging task of single-view future prediction by leveraging full 3D reconstruction from partial inputs. The table also compares our results without inverse physics-based hand refinement, where our refinement is shown to be significantly effective when the input view is highly sparse to sufficiently constrain hand model fitting.

\vspace{-0.2\baselineskip}
To further evaluate hand fitting accuracy, we additionally report the metric $\textrm{Hand CD}$ in the table, which measures the Chamfer Distance between the fitted hand meshes and the ground-truth 3D hand point cloud lifted from the multi-view depth maps available in the dataset.\footnote{Note that in other multi-view experiments, these multi-view depth maps are used as \emph{inputs} during training and are therefore not treated as ground truth for evaluation. We kindly refer the reader to our supplementary for detailed discussion.} Our hand refinement also noticeably improves this hand fitting accuracy metric; to the best of our knowledge, this is the first work to demonstrate that a physics-based deformable object prior can benefit hand reconstruction.

\input{tab/single_view_dh}

\vspace{-0.5\baselineskip}
\subsection{Robustness Analysis}
\vspace{-0.2\baselineskip}
\label{subsec:robustness_analysis}

We also compare the robustness of our method and PhysTwin~\citep{jiang2025phystwin} under perturbed input signals, including input depth, CoTracker~\citep{karaev2024cotracker3} tracking results, and MANO~\citep{romero2017embodied}-based hand fitting results. 
Specifically, we add ~1 mm noise to the input depth, ~1 px perturbation to the CoTracker tracks, and perturbations to the MANO parameters such that the resulting hand pose has an MPJPE of ~10 mm. 

In Tab.~\ref{tab:noise_all}, PhysTwin exhibits larger performance degradation across all settings, particularly under perturbed tracking signals. In contrast, our method shows noticeably smaller performance drops. This robustness comes from using denser hand reconstruction, which provides more reliable contact cues and mitigates the impact of upstream noise; as a result, perturbations to depth, tracking, or controller parameters lead to only modest accuracy changes. These results highlight the potential of our method to scale to real-world applications using monocular or multi-view RGB-only videos, where our model would be fitted to depth or hand estimates \textit{predicted} from RGB, since our simulated noise levels remain within the accuracy range of current RGB-based state-of-the-art estimators.


\vspace{-0.2\baselineskip}
\input{tab/noise_all_cm}

\vspace{-1\baselineskip}
\subsection{Contact Topology Analysis}
\vspace{-0.2\baselineskip}
\label{subsec:contact_analysis}

\begin{figure*}[!h]
\begin{center}
\includegraphics[width=0.95\textwidth]{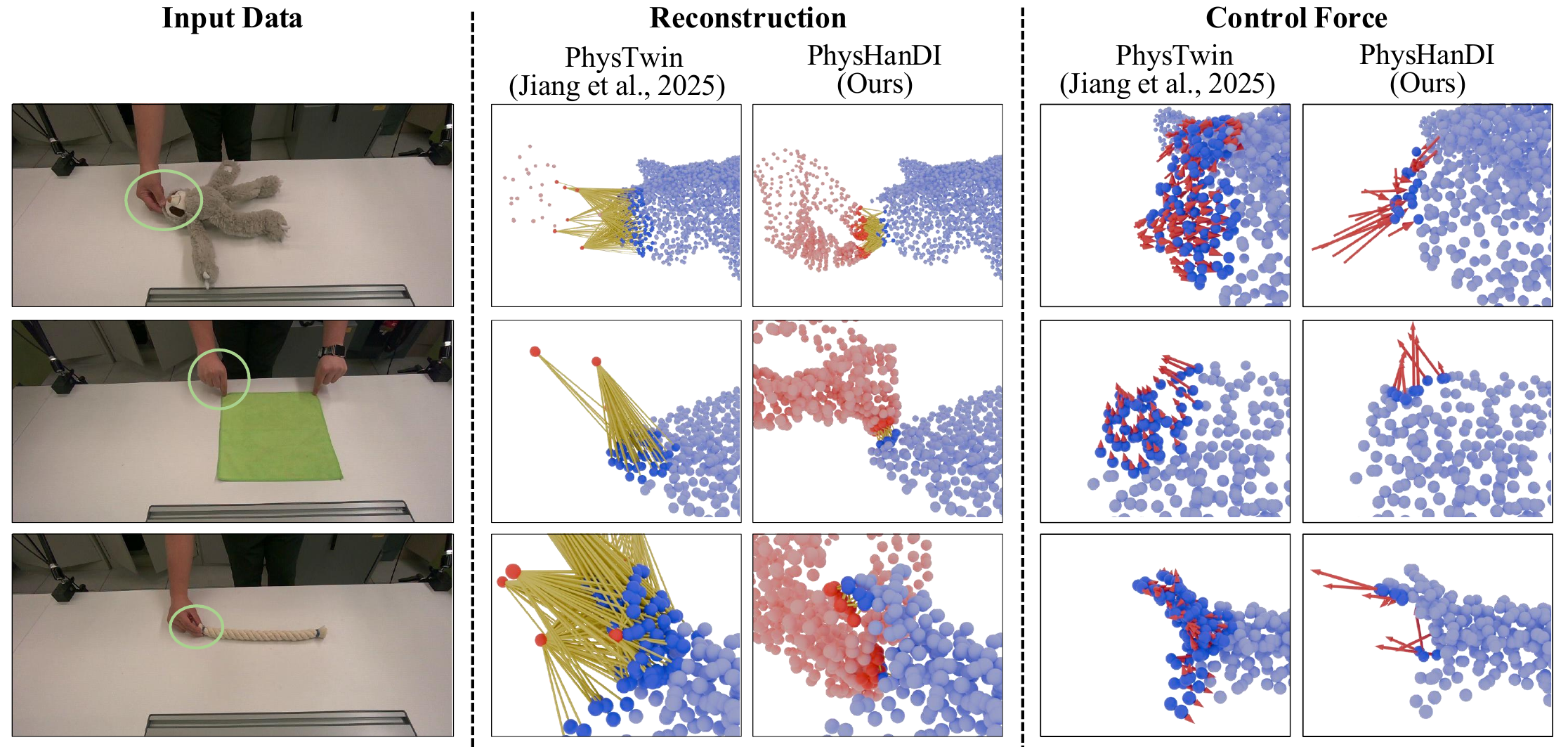}
\vspace{-0.5\baselineskip}
\caption{\textbf{Comparisons in reconstructed spring–mass model topology and force.} (1) \textit{Topology reconstruction}. PhysTwin~\citep{jiang2025phystwin}'s sparser hand points tend to result in excessively long virtual spring lengths to maintain contact coverage, whereas ours based on dense hand points precisely localizes contacts without unnecessary spring elongation—considered a more optimal topology in prior works~\citep{silling2005meshfree,silling2007peridynamic,wang2023determination}. (2) \emph{Control force visualization.} PhysTwin's broader spring coverage disperses forces across non-contact regions, whereas ours concentrates forces at the actual contact, favoring local, detailed manipulation. \emph{Visualization key}. Yellow line segments depict virtual-springs $\mathcal{E}^{\text{virtual}}$. Blue spheres denote object nodes $\mathcal{N}$ and red spheres denote control nodes $\mathcal{V}'$. Red arrows visualize control-force vectors induced on object nodes by virtual springs.}
\label{fig:contact}
\end{center}
\vspace{-0.4\baselineskip}
\end{figure*}

In this section, we further discuss why our object simulation achieves better performance than the current state-of-the-art method, PhysTwin~\citep{jiang2025phystwin}, as empirically shown in the previous subsections.
Specifically, we examine differences in the optimized spring-mass model topology, since \textit{``the behavior of the [spring-mass] model is dependent on the topology''}~\citep{nealen2006physically}.

As discussed in Sec.~\ref{sec:method}, the connection radius $\delta$ is the primary factor defining the model topology in both our method and PhysTwin, thus we present a numerical analysis of the fitted $\delta$.
In particular, we refer to practical analyses of particle-based models with radius-limited neighbor interactions (e.g., peridynamics)~\citep{silling2005meshfree,silling2007peridynamic,wang2023determination}, as our spring-mass model likewise restricts interactions to neighbors within a distance cutoff $\delta$.
They show that, given the spatial discretization resolution $\Delta x$ of the object\footnote{In our case, $\Delta x$ is approximated as the mean distance to each node's four nearest neighbors, averaged across all nodes.}, the ratio $\delta / \Delta x$ should remain close to a small constant $r$, and that \textit{``values much larger than this may result in excessive wave dispersion and require very large computer run times.''}~\citep{silling2005meshfree}

Inspired by this, we introduce a simple measure of deviation from this recommended ratio of connection radius to discretization resolution.
Specifically, we report the \textbf{Radius-to-Resolution Deviation ($RRD$)},
\vspace{-0.2\baselineskip}
\begin{equation}
    RRD = |(\delta / \Delta x)/r - 1|,
\end{equation}
\vspace{-0.6\baselineskip}

\noindent where we use $r=3$ as the reference value, reflecting a value commonly acknowledged as plausible in the prior works~\citep{silling2005meshfree,silling2007peridynamic,wang2023determination}.
As shown in Tab.~\ref{tab:radius_analysis}, our method yields about 2$\times$ lower $RRD$ for object springs and over 7$\times$ lower $RRD$ for virtual springs compared to PhysTwin~\citep{jiang2025phystwin}, indicating that our model topology is more optimal according to the analyses in the aforementioned literature.

Related to these results, we also show the topology and contact visualization in Fig.~\ref{fig:contact}, where PhysTwin's sparser control points are likely to result in excessively long virtual-spring lengths to maintain contact coverage, whereas our denser hand reconstruction precisely localizes contacts without unnecessarily elongating the springs.
In addition, the control force visualization (right column of Fig.~\ref{fig:contact}) shows that PhysTwin's wider virtual-spring coverage diffuses forces over a larger area, weakening local actuation around the true contact. In contrast, ours concentrates forces only where contact actually occurs, which is preferable for fine manipulation.
These analyses suggest that dense hand reconstruction improves topology optimization of the spring--mass model, yielding a smaller, resolution-matched connection radius and more reliable dynamics.

\input{tab/radius_analysis}

%% file: tab/single_view_dh.tex
\begin{table}[!h]
\centering
\small
\resizebox{\columnwidth}{!}{%
\begin{tabular}{c|ccc|cc}
\toprule
\multirow{2}{*}{\textbf{Method}} & \multicolumn{3}{c|}{\textbf{3D Metrics}} & \multicolumn{2}{c}{\textbf{2D Metrics}} \\
& CD $\downarrow$ & Track Err. $\downarrow$ & Hand CD $\downarrow$ & IoU $\uparrow$ & PSNR $\uparrow$ \\
\midrule
\textbf{\textsc{PhysHanDI} (Ours) - Hand Ref.} & 42.8 & 7.36 & 7.57 & 0.49 &	19.50 \\
\textbf{\textsc{PhysHanDI} (Ours)} & \textbf{33.5} & \textbf{6.75} & \textbf{7.17}  & \textbf{0.51} &	\textbf{19.67} \\
\bottomrule
\end{tabular}
}
\vspace{0.5\baselineskip}

\caption{\textbf{Single-view future prediction results on the PhysTwin-full dataset~\citep{jiang2025phystwin}.} Notably, our hand refinement using the physics-based object prior is effective in enhancing hand reconstruction quality.}

\label{tab:single_view_future_pred}
\vspace{-2\baselineskip}
\end{table}

%% file: tab/noise_all_cm.tex
\begin{table}[!h]
\centering
\fontsize{7pt}{7pt}\selectfont
\setlength{\tabcolsep}{1pt}
\renewcommand{\arraystretch}{1.1}

\begin{adjustbox}{width=\linewidth}
\begin{tabularx}{\linewidth}{>{\centering\arraybackslash}m{1.7cm}|YY|YY}
\toprule
\multirow{2}{*}{\textbf{Method}} &
\multicolumn{2}{c|}{\textbf{PhysTwin-dense}} &
\multicolumn{2}{c}{\textbf{PhysTwin-full}} \\
\cmidrule(lr){2-3}\cmidrule(lr){4-5}
& CD $\downarrow$ & Track Err. $\downarrow$ & CD $\downarrow$ & Track Err. $\downarrow$ \\
\midrule
\multicolumn{5}{c}{\textbf{(a) Clean Input}}\\
\midrule
PhysTwin & 5.90 & 1.00 & 5.52 & 0.97 \\
\textbf{Ours} & \textbf{5.30} & \textbf{0.89} & \textbf{5.40} & \textbf{0.96} \\
\midrule
\multicolumn{5}{c}{\textbf{(b) Perturbed Depth}}\\
\midrule
PhysTwin & 6.93 (1.03) & 1.12 (0.12) & 6.34 (0.82) & 1.12 (0.15) \\
\textbf{Ours} & \textbf{6.19} (\textbf{0.89}) & \textbf{1.00} (\textbf{0.11})
      & \textbf{6.10} (\textbf{0.70}) & \textbf{1.05} (\textbf{0.09}) \\
\midrule
\multicolumn{5}{c}{\textbf{(c) Perturbed Tracking Signal}}\\
\midrule
PhysTwin & 9.60 (3.70) & 1.46 (0.46) & 8.32 (2.80) & 1.34 (0.37) \\
\textbf{Ours} & \textbf{5.56} (\textbf{0.26}) & \textbf{0.86} (\textbf{-0.03})
      & \textbf{5.50} (\textbf{0.10}) & \textbf{0.92} (\textbf{-0.04}) \\
\midrule
\multicolumn{5}{c}{\textbf{(d) Perturbed Controller}}\\
\midrule
PhysTwin & 7.54 (1.64) & 1.25 (0.25) & 6.89 (\textbf{1.37}) & \textbf{1.25} (\textbf{0.28}) \\
\textbf{Ours} & \textbf{6.44} (\textbf{1.14}) & \textbf{1.08} (\textbf{0.19})
           & \textbf{6.79} (1.39) & 1.29 (0.33) \\
\bottomrule
\end{tabularx}
\end{adjustbox}

\vspace{0.5\baselineskip}

\caption{\textbf{Robustness analysis} under perturbations applied to depth input, CoTracker~\citep{karaev2024cotracker3} trajectories, and hand-pose controller parameters. We compare PhysTwin~\citep{jiang2025phystwin} and our method under the same settings. Values in parentheses denote performance changes relative to the clean-input baseline.}

\vspace{-1.2\baselineskip}
\label{tab:noise_all}
\end{table}

%% file: tab/radius_analysis.tex
\vspace{-0.5\baselineskip}
\begin{table}[!h]
\centering
\small
\resizebox{1.0\linewidth}{!}{%
\begin{tabular}{c|cc}
\toprule
Method & $RRD_\text{object}$ $\downarrow$ & $RRD_\text{virtual}$ $\downarrow$  \\
\midrule
PhysTwin~\citep{jiang2025phystwin} & 0.64 & 2.63 \\
\textbf{\textsc{PhysHanDI}} (Ours) & \textbf{0.32} & \textbf{0.35} \\
\bottomrule
\end{tabular}
}
\vspace{0.5\baselineskip}
\caption{\textbf{Radius-to-Resolution Deviation ($RRD$)} for object and virtual springs (lower is better). $RRD = |(\delta / \Delta x)/r - 1|$. Our method achieves about 2$\times$ lower $RRD$ for object springs and over 7$\times$ lower for virtual springs compared to PhysTwin~\citep{jiang2025phystwin}.}
\vspace{-\baselineskip}
\label{tab:radius_analysis}
\end{table}

%% file: sec/06_conclusion.tex
\vspace{-1.2\baselineskip}

\section{Conclusion}
\label{sec:conclusion}
\vspace{-0.2\baselineskip}

We presented \textsc{PhysHanDI}, a physics-based framework for modeling and reconstructing hand–object interactions involving highly non-rigid objects. By incorporating physical priors and simulating object deformations driven by forces from the fully reconstructed 3D hands, our method produces reconstructions that are both physically plausible and consistent with interacting hand dynamics. Through a reconstruction pipeline based on sparse-view RGB-D inputs, \textsc{PhysHanDI} demonstrates superior performance over existing baselines in reconstruction, future prediction, and generalization to unseen interactions. This work takes a step toward more general and robust modeling of everyday hand–object interactions, opening up new opportunities for applications in embodied AI and digital human modeling.

%

%% file: sec/07_appendix.tex
\section{Dataset Details}
\label{sec:dataset_details}

\begin{figure*}[!h]
\begin{center}
\includegraphics[width=\textwidth]{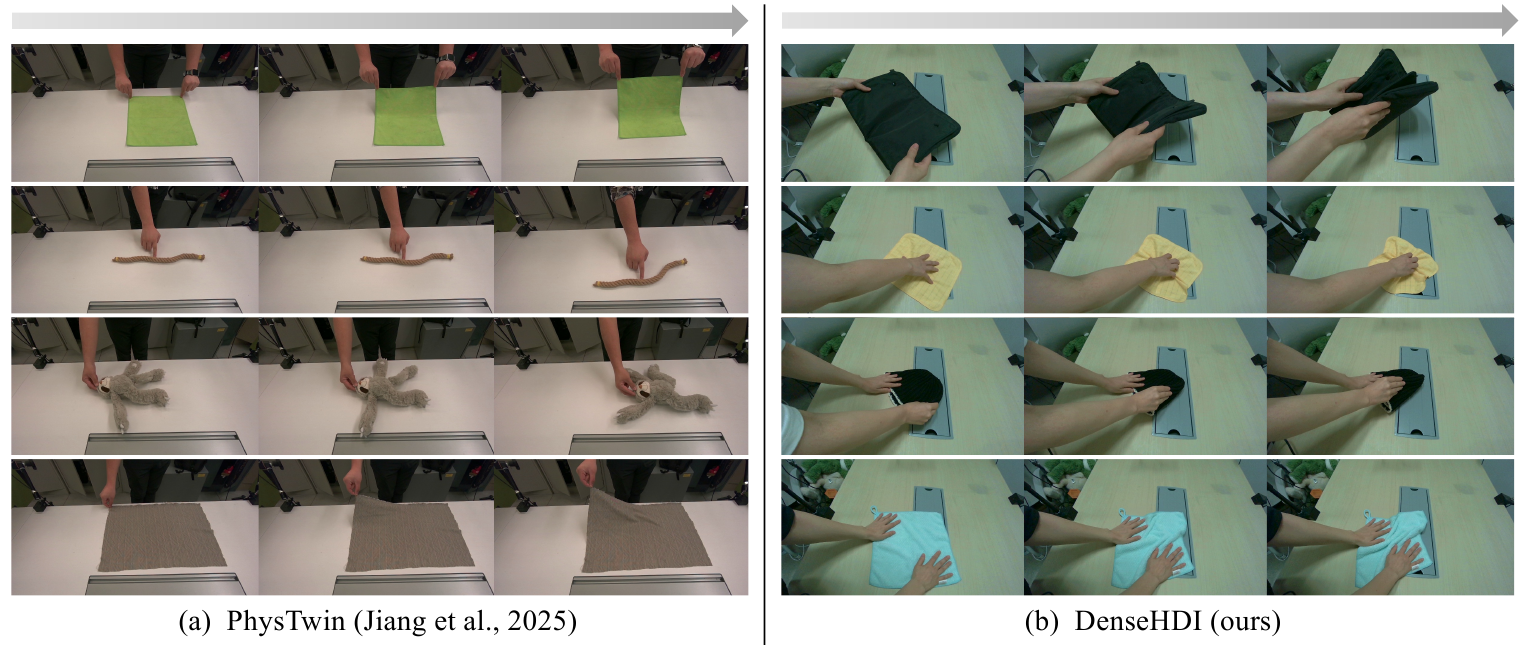}
\vspace{-\baselineskip}
\caption{\textbf{Captured sequences in PhysTwin~\citep{jiang2025phystwin} and \textsc{DenseHDI}}. The sequences in \textsc{DenseHDI} feature denser hand–object contacts.}
\label{fig:dataset_comparisons}
\end{center}
\vspace{-0.5\baselineskip}
\end{figure*}

In this section, we present the details of our newly captured dataset, \textsc{DenseHDI}, introduced in Sec.~\ref{subsec:experiment_settings}.
For data acquisition and pre-processing, we follow the same protocol as PhysTwin~\citep{jiang2025phystwin}, using three RealSense D455 RGB-D cameras to record three-view videos of hand–deformable object interactions.
In total, we collect 19 sequences, each lasting 2–8 seconds, spanning 10 object types (e.g., swimming cap, cloth, pouch, towel). The dataset includes diverse interactions, such as folding a pouch or towel and squeezing a cloth.
We note that the existing PhysTwin dataset~\citep{jiang2025phystwin} primarily captures sparse, point-like hand–object contacts (e.g., pointing at or pushing with one finger, or pinching with two fingers), whereas our dataset focuses on capturing denser hand–object contacts, such as wiping with a dishcloth or folding a pouch using the palm.
Visualizations of these captured sequences are provided in Fig.~\ref{fig:dataset_comparisons}.

\section{Method Details}
\label{sec:method_details}

In this section, we provide additional details on reconstructing our dense hand–deformable object interaction model from sparse-view RGB-D video inputs, as discussed in Sec.~\ref{subsec:handi_training}.

\subsection{Hand Reconstruction}
In the \textit{hand reconstruction} stage, we fit the MANO model~\citep{romero2017embodied} to the input multi-view RGB-D videos using the loss function defined in Eq.~\ref{eq:hand_recon_loss} (Sec.~\ref{subsec:handi_training}).
$\mathcal{L}_{\textit{2D}}$, $\mathcal{L}_{\textit{d}}$, and $\mathcal{L}_{\textit{t}}$ are defined as L2 losses, with $\lambda_{\textit{d}}$ and $\lambda_{\textit{t}}$ set to $1 \times 10^2$ and $5 \times 10^{5}$, respectively.

To obtain 2D keypoint supervision for computing $\mathcal{L}_{\textit{2D}}$, we use an off-the-shelf estimator (MediaPipe~\citep{zhang2020mediapipe}).
However, we empirically observe that it yields missing or implausible predictions for heavily occluded hand joints, which degrade the MANO fitting results—particularly in our sparse-view setting.
To mitigate this, we additionally obtain 2D keypoint supervision $\mathbf{U}^{\text{mano}}$ from a \emph{monocular MANO parameter estimator}~\citep{dong2024hamba}, which yields plausible predictions constrained by the MANO space, though with less precise 2D alignment.\footnote{See related discussions in prior works, e.g., \cite{li2021hybrik}. Although the MANO-based estimator predicts full 3D hand shapes and poses, we use only its 2D projections since its depth estimates are ambiguous due to the \emph{monocular} setting (e.g., projective ambiguity, scale–depth trade-off).} 
We empirically find that combining discrepancy losses with respect to $\mathbf{U}^{2D}$ (from MediaPipe~\citep{zhang2020mediapipe}) and $\mathbf{U}^{\text{mano}}$ yields more robust MANO fitting, with the loss weight for $\mathbf{U}^{\text{mano}}$ set to 0.5.

For optimizing the MANO parameters based on the aforementioned loss, we use the AdamW optimizer~\citep{loshchilov2017decoupled} for 1500 steps with a learning rate of $2 \times 10^{-3}$, decaying by a factor of 0.98 every 40 steps. The MANO parameters at each frame are initialized from the fitting results of the previous frame, while the first frame is initialized randomly.

\subsection{Object Reconstruction}
\label{subsec:obj_recon}

After the hand reconstruction stage, we fit the spring–mass model~\citep{liu2013fast,jiang2025phystwin} representing the deformable object, conditioned on the previously fitted 3D hands.
Directly following \cite{jiang2025phystwin}, we adopt a hierarchical optimization scheme with (1) a \textbf{sparse (zero-order)} stage followed by (2) a \textbf{dense (first-order)} stage.

\keyword{Sparse (zero-order) stage.}
We optimize the coarse, non-differentiable spring–mass model parameters $\Theta_{0} = \{\mathcal{T}, s_{\text{global}}, \eta\}$, where $\mathcal{T}$ denotes the spring–mass topology parameterized by a connection radius $\delta$ and a maximum number of connected nodes $d_{\text{max}}$, $s_{\text{global}}$ is the global spring stiffness (assuming homogeneity at this stage), and $\eta$ represents collision parameters.

\keyword{Dense (first-order) stage.}
With $\Theta_{0}$ fixed, we refine the differentiable per-spring parameters $\Theta_{1} = \{s_{ij}, \gamma_{ij}\}_{(i, j) \in \mathcal{E} \cup \mathcal{E}^{\text{virtual}}}$, where $s_{ij}$ and $\gamma_{ij}$ denote per-spring stiffness and damping parameters.

The optimization objective at each stage $k \in \{0, 1\}$ (where $k=0$ and $k=1$ correspond to the sparse and dense stages, respectively) can be written as:

\begin{equation}
\min_{\Theta_{\textit{k}}}
\;\; \frac{1}{T} \sum_{t=1}^T \mathcal{L}_{\textit{ch}}(\hat{\mathbf{S}}_t, \mathbf{S}_t)
+ \lambda 
\mathcal{L}_{\textit{tr}}(\hat{\mathbf{S}}_t, \mathbf{S}_t) \quad
\text{s.t.} \quad \hat{\mathbf{S}}_{t} = f(\hat{\mathbf{S}}_{t-1}, \Theta_{0}, \Theta_{1}, \Theta_{\textit{h}}),
\, \hat{\mathbf{S}}_{0} = {\mathbf{S}}_{0},
\label{eq:loss_function}
\end{equation}
\noindent Here, $\hat{\mathbf{S}}_{t}$ denotes the simulated state, ${\mathbf{S}}_{t}$ denote the observed state from the inputs, and $f$ denote a simulation forward function based on the spring-mass system.
As discussed in Sec.~\ref{subsec:handi_training}, $\mathcal{L}_{\textit{ch}}$ measures the Chamfer distance to encourage simulated nodes to remain close to the 3D point cloud lifted from the input depth maps, while $\mathcal{L}_{\textit{tr}}$ measures the $\ell_2$ discrepancy to per-frame 3D tracked points obtained from CoTracker3~\citep{karaev2024cotracker3}.

For the sparse stage, we use zero-order optimization~\citep{lozano2006towards} for 100 iterations, with initialization values of $\delta$, $d_{\text{max}}$, and $s_{\text{global}}$ set to 0.002 and 3, respectively.
For the dense stage, we use the Adam optimizer~\citep{kingma2014adam} for 200 iterations with an initial learning rate of $1 \times 10^{-3}$. All other hyperparameters are kept identical to \cite{jiang2025phystwin}.

\subsection{Hand Refinement}
In this stage, we refine the initial MANO parameters $\Theta_{\textit{h}}$ to produce object simulations better aligned with the input observations, using the spring–mass model fitted in the previous stage.
An overview of this stage, including the loss function, is provided in Sec.~\ref{subsec:handi_training}. For the loss function in Eq.~\ref{eq:hand_recon_loss}, we set $\lambda_{\textit{tr}} = 1$.
Optimization is performed with the Adam optimizer~\citep{kingma2014adam} with 40 optimization steps. The MANO parameters are initialized from the fitting results of the initial hand reconstruction stage, with an initial learning rate of $2 \times 10^{-5}$ decayed by 0.99 at each iteration.


\section{Quantitative Results on the PhysTwin-full Dataset}
Tab.~\ref{tab:recon_future_phystwin_full_dataset} reports quantitative results on the PhysTwin-full dataset~\cite{jiang2025phystwin}. As discussed in Sec.~\ref{subsec:experiment_settings}, most sequences in this dataset involve sparse point-based hand–object contacts, which are less representative of realistic interactions but can favor PhysTwin due to its reliance on a sparse point-based controller. In this setting, our method still mostly ourperforms the baselines~\cite{jiang2025phystwin,zhong2024reconstruction,zhang2024dynamics}. 

\input{tab/appendix_comp_dynamic_CD_dh}

\vspace{0.5\baselineskip}
We additionally provide a per-sequence breakdown on representative sequences from the PhysTwin-full dataset in Tab.~\ref{tab:per_sequence_breakdown}, where the sequences are categorized into \emph{dense}- and \emph{sparse}-contact groups. 
As shown in the table, our method consistently improves over PhysTwin across all sequences, with substantially larger gains on dense-contact sequences (e.g., $\text{CD}_\text{dyn}$ of 6.23 vs.\ 21.30 on \texttt{double\_stretch\_sloth}). This per-sequence analysis confirms that the prior state of the art, PhysTwin~\citep{jiang2025phystwin}, performs well for sparse point-based hand-object contacts due to its sparse controller representation, but is less robust to dense contacts. In contrast, our method remains robust in both sparse- and dense-contact settings.

\input{tab/per_sequence_breakdown}

\vspace{-0.5\baselineskip}

\section{Generalization to Unseen Interactions}
\label{subsec:gen_unseen_inter}

In this section, we evaluate reconstruction quality on novel interaction sequences performed on the same object but with different interaction types, following the evaluation protocol of \citep{jiang2025phystwin}.
As shown in Tab.~\ref{tab:unseen_inter_dh}, our method outperforms PhysTwin on all metrics, demonstrating strong generalizability to unseen interaction actions.


\input{tab/unseen_inter_dh}

\section{Contact Consistency Analysis}
\label{sec:contact_consistency}

In this section, we provide a quantitative evaluation of contact consistency in the reconstructed hand--deformable-object interactions.
We follow common evaluation protocols from prior hand--rigid-object interaction works~\citep{grady2021contactopt,liu2023contactgen} and construct \emph{pseudo contact labels} based on the spatial proximity between object and hand points, using distance thresholds of $d < 5$\,mm and $10$\,mm.
We then measure \textbf{Contact Accuracy} as the agreement rate between the predicted hand--object contacts and these pseudo labels.

As shown in Tab.~\ref{tab:contact_consistency}, our method achieves higher Contact Accuracy than PhysTwin at both the 5\,mm and 10\,mm contact distance thresholds, indicating more consistent contact estimation. This observation aligns with our qualitative results in Fig.~\ref{fig:contact} and Sec.~\ref{subsec:contact_analysis}, where dense hand reconstruction enables more accurately localized virtual-spring connections at true contact regions.

\input{tab/contact_consistency}



\section{Sensitivity to Initial Hyperparameters}
\label{sec:hyperparam_initial}

We additionally analyze the sensitivity of our method to the initial hyperparameters used in the zero-order optimization during the object reconstruction stage (Sec.~\ref{subsec:obj_recon}): the initial connection radius $\delta$ and the maximum number of connected nodes $d_\text{max}$. Tab.~\ref{tab:hyperparam_initial} reports the results, where row B corresponds to our default setting ($\delta=0.002$, $d_\text{max}=3$). Rows C--E and F--H correspond to settings with varying $\delta$ and $d_\text{max}$, respectively. We observe that our method remains robust within a reasonable range of initial values.


\input{tab/hyperparam_initial}

\section{Computational Cost Comparison}
\label{sec:computational_cost}

We provide a detailed runtime breakdown of our method and PhysTwin~\citep{jiang2025phystwin}. Tab.~\ref{tab:timecost} reports the average per-frame runtime of each stage in both pipelines. Our method introduces additional computation time for hand reconstruction and refinement, which PhysTwin does not perform but which are essential for accurate full 3D hand modeling and inverse-physics-based hand refinement that benefits both hand and object reconstruction (Sec.~\ref{subsec:future_pred}). In contrast, our object reconstruction stages and inference-time simulation are slightly faster than the corresponding stages of PhysTwin. This is related to our spring-mass topology analysis in Sec.~\ref{subsec:contact_analysis}: PhysTwin's sparse control points require an excessively large connection radius to maintain contact coverage, which prior literature notes can result in \textit{``excessive wave dispersion and require very large computer run times''}~\citep{silling2005meshfree}.

\input{tab/timecost}

\section{Discussions \& Limitations}

\keyword{Evaluation of multi-view hand reconstruction.}
Although we directly evaluated hand reconstruction accuracy in the single-view setting (Sec.~\ref{subsec:future_pred}) using ground-truth hand point clouds lifted from multi-view depth maps, this evaluation is not possible in the main multi-view experiments since those depth maps are used as inputs during training and more precise hand annotations are unavailable.
Indeed, in prior work (e.g., \citep{hampali2020honnotate}), MANO fitting to multi-view RGB-D is often treated as a way to \emph{annotate ground-truth 3D hand meshes} in datasets lacking such labels, and fitting quality is typically assessed indirectly via downstream applications.
Motivated by this, we evaluate hand fitting quality primarily through physics-based deformable object reconstruction, where more accurate hands directly yield more accurate object simulations. Nonetheless, a direct evaluation would be valuable—for example, by capturing \emph{denser-view ground-truth} 3D hands and comparing them against our sparse multi-view reconstructions, if such a capture system is available.

\keyword{Handling dynamic contact changes.}
As discussed in Sec.~\ref{subsec:representation}, our interaction force is modeled to encourage the maintenance of the contact topology (i.e., the rest length of the virtual springs between hand vertices and object nodes). This modeling assumes that the hand–object contact topology remains static within a sequence. While this assumption is also common in existing hand-rigid-object interaction reconstruction methods~\cite{hampali2020honnotate,cho2024dense}, handling dynamic hand–object contact changes would be an important direction for future research.
%
%
In addition, such interaction force modeling does not account for the actual force (e.g., finger pressure) but instead serves as a boundary condition to drive the simulation of the spring–mass model. Explicitly modeling the actual hand force would be non-trivial, yet an interesting future research direction with potential applications in haptics.

%% file: tab/appendix_comp_dynamic_CD_dh.tex
\begin{table}[!h]
\centering
\small
\resizebox{\linewidth}{!}{%
\begin{tabular}{c|ccc|cc|ccc|cc}
\toprule
\multirow{3}{*}{\textbf{Method}} & \multicolumn{5}{c|}{\textbf{Reconstruction \& Resimulation}} & \multicolumn{5}{c}{\textbf{Future Prediction}} \\
\cmidrule(lr){2-11}
\multirow{2}{*}{} & \multicolumn{3}{c|}{\textbf{3D Metrics}} & \multicolumn{2}{c|}{\textbf{2D Metrics}} & \multicolumn{3}{c|}{\textbf{3D Metrics}} & \multicolumn{2}{c}{\textbf{2D Metrics}} \\
 & $\text{CD}_\text{dyn}$ $\downarrow$ & $\text{CD}_\text{full}$ $\downarrow$ & Track Err. $\downarrow$ & IoU $\uparrow$ & PSNR $\uparrow$ & $\text{CD}_\text{dyn}$ $\downarrow$ & $\text{CD}_\text{full}$ $\downarrow$ & Track Err. $\downarrow$ & IoU $\uparrow$ & PSNR $\uparrow$ \\
\midrule
Spring-Gaus~\cite{zhong2024reconstruction} &	26.39 & 33.60& 4.07&	0.62&	21.24&	49.29 &	46.54& 6.61 &	0.48&	19.59 \\
GS-Dynamics~\cite{zhang2024dynamics} & 24.73 & 13.79& 2.18& 0.72& 24.01& 52.96 & 38.84& 6.88& 0.46& 19.38\\
PhysTwin~\cite{jiang2025phystwin} & 7.63& 5.52&	0.97&	\textbf{0.84}&	26.32& 14.42&	12.26&	2.44&	\textbf{0.69}&	22.80 \\
\textbf{\textsc{PhysHanDI} (Ours)} &	\textbf{7.30}& \textbf{5.40}&  \textbf{0.96}& \textbf{0.84}&	\textbf{26.44}&	\textbf{13.63}&	\textbf{12.04}&  \textbf{2.41}&	0.68&	\textbf{22.96}\\

\bottomrule
\end{tabular}
}
\vspace{0.5\baselineskip}
\caption{\textbf{Reconstruction \& Resimulation and Future Prediction results on the PhysTwin-full dataset~\citep{jiang2025phystwin}}. Our method outperforms the state-of-the-art~\citep{jiang2025phystwin} on most metrics. $\textrm{CD}$ is measured in millimeters, and $\textrm{Track Err.}$ is scaled by $\times100$ for readability.}
\label{tab:recon_future_phystwin_full_dataset}
\vspace{-1.5\baselineskip}
\end{table}

%% file: tab/per_sequence_breakdown.tex
\begin{table}[!h]
\centering
\small
\begin{tabular}{llcc}
\toprule
\textbf{Contact Type} & \textbf{Sequence} & \textbf{PhysTwin}~\cite{jiang2025phystwin} & \textbf{\textsc{PhysHanDI} (Ours)} \\
\midrule
\multirow{3}{*}{Dense}  & \texttt{double\_lift\_cloth\_3}   & 12.21 & \textbf{6.70} \\
                        & \texttt{double\_lift\_sloth}      & 5.33  & \textbf{4.39} \\
                        & \texttt{double\_stretch\_sloth}   & 21.30 & \textbf{6.23} \\
\midrule
\multirow{3}{*}{Sparse} & \texttt{single\_lift\_cloth\_1}   & 10.55 & \textbf{10.47} \\
                        & \texttt{single\_lift\_cloth\_4}   & 6.73  & \textbf{5.87} \\
                        & \texttt{single\_push\_rope}       & 3.81  & \textbf{3.61} \\
\bottomrule
\end{tabular}
\vspace{0.5\baselineskip}
\caption{\textbf{Per-sequence $\text{CD}_\text{dyn}$ comparison on representative sequences from the PhysTwin-full dataset~\citep{jiang2025phystwin}.} Results are reported for reconstruction and resimulation with multi-view RGB-D inputs. $\text{CD}_\text{dyn}$ is measured in millimeters. }
\label{tab:per_sequence_breakdown}
\vspace{-1.0\baselineskip}
\end{table}

%% file: tab/unseen_inter_dh.tex
\begin{table}[!h]
\centering
\small
\resizebox{0.6\columnwidth}{!}{%
\begin{tabular}{c|cc|cc}
\toprule
\multirow{2}{*}{\textbf{Method}} & \multicolumn{2}{c|}{\textbf{3D Metrics}} & \multicolumn{2}{c}{\textbf{2D Metrics}} \\
& CD $\downarrow$ & Track Err. $\downarrow$ & IoU $\uparrow$ & PSNR $\uparrow$ \\
\midrule
PhysTwin~\citep{jiang2025phystwin} & 8.94 & 1.77 & 0.79 &	25.44 \\
\textbf{\textsc{PhysHanDI}} (Ours) & \textbf{8.38} & \textbf{1.70} & \textbf{0.82} &	\textbf{25.89} \\
\bottomrule
\end{tabular}
}
\vspace{0.6\baselineskip}
\caption{\textbf{Generalization to unseen interactions on the PhysTwin-full dataset~\citep{jiang2025phystwin}.} Our method demonstrates superior generalizability compared to the state of the art~\citep{jiang2025phystwin}.}
\label{tab:unseen_inter_dh}
\vspace{-2\baselineskip}

\end{table}

%% file: tab/contact_consistency.tex
\begin{table}[!h]
\centering
\small
\begin{tabular}{lcc}
\toprule
\textbf{Method} & \textbf{Acc.@5\,mm (\%)} $\uparrow$ & \textbf{Acc.@10\,mm (\%)} $\uparrow$ \\
\midrule
PhysTwin~\cite{jiang2025phystwin} & 97.0 & 97.5 \\
\textbf{\textsc{PhysHanDI} (Ours)} & \textbf{98.2} & \textbf{98.3} \\
\bottomrule
\end{tabular}
\vspace{0.5\baselineskip}
\caption{\textbf{Quantitative comparison on contact consistency.} Contact Accuracy (\%) is computed against pseudo contact labels constructed from the spatial proximity between object and hand points, with distance thresholds of 5\,mm and 10\,mm—following protocols similar in spirit to~\citep{grady2021contactopt,liu2023contactgen}. Our method achieves higher accuracy than PhysTwin~\citep{jiang2025phystwin} at both thresholds, indicating more precise and localized contact estimation.}
\label{tab:contact_consistency}
\vspace{-1.0\baselineskip}
\end{table}

%% file: tab/hyperparam_initial.tex
\begin{table}[!h]
\centering
\small
\begin{tabular}{clcccc}
\toprule
\textbf{Row} & \textbf{Method} & \textbf{Initial $\delta$} & \textbf{Initial $d_\text{max}$} & $\text{CD}_\text{full}$ $\downarrow$ & \textbf{PSNR} $\uparrow$ \\
\midrule
A & PhysTwin~\cite{jiang2025phystwin}                    & 0.040   & 50  & 8.86          & 22.48 \\
B & \textbf{\textsc{PhysHanDI} (Ours, default)}          & 0.002  & 3   & \textbf{4.44} & \textbf{24.60} \\
\midrule
C & \textsc{PhysHanDI} (Ours)                            & 0.001  & 3   & 6.81          & 22.62 \\
D & \textsc{PhysHanDI} (Ours)                            & 0.020   & 3   & 4.62          & 24.18 \\
E & \textsc{PhysHanDI} (Ours)                            & 0.040   & 3   & 5.26          & 23.98 \\
\midrule
F & \textsc{PhysHanDI} (Ours)                            & 0.002  & 1   & \underline{4.47}          & \underline{24.24} \\
G & \textsc{PhysHanDI} (Ours)                            & 0.002  & 10  & 4.90          & 24.09 \\
H & \textsc{PhysHanDI} (Ours)                            & 0.002  & 50  & 6.15          & 23.59 \\
\bottomrule
\end{tabular}
\vspace{0.5\baselineskip}
\caption{\textbf{Sensitivity to initial hyperparameters} used in the zero-order optimization of the object reconstruction stage on the \texttt{double\_stretch\_sloth} sequence (reconstruction and resimulation with multi-view RGB-D inputs). Row B is our default setting. Rows C--E vary the initial connection radius $\delta$ while fixing $d_\text{max}=3$; rows F--H vary the initial maximum number of connected nodes $d_\text{max}$ while fixing $\delta=0.002$. Our method remains robust within a reasonable range of initial values.}\label{tab:hyperparam_initial}
\vspace{-0.5\baselineskip}
\end{table}

%% file: tab/timecost.tex
\begin{table}[!h]
\centering
\small
\resizebox{\linewidth}{!}{%
\begin{tabular}{l|cccc|c}
\toprule
\multirow{2}{*}{\textbf{Method}} & \multicolumn{4}{c|}{\textbf{Training}} & \multirow{2}{*}{\textbf{Inference}} \\
\cmidrule(lr){2-5}
 & Hand Recon. & Object Recon. (Zero-order) & Object Recon. (First-order) & Hand Refine. & \\
\midrule
PhysTwin~\cite{jiang2025phystwin}    & --     & 13.80          & 21.22          & --     & 0.14 \\
\textbf{\textsc{PhysHanDI} (Ours)}   & 27.38  & \textbf{12.24} & \textbf{17.63} & 12.32  & \textbf{0.11} \\
\bottomrule
\end{tabular}
}
\vspace{0.5\baselineskip}
\caption{\textbf{Average per-frame runtime breakdown (in seconds) of our method and PhysTwin~\citep{jiang2025phystwin}.} Our object reconstruction stages, including zero-order and first-order optimization, and inference-time simulation are slightly faster than the corresponding stages of PhysTwin. The remaining overhead in our pipeline comes from the additional Hand Reconstruction and Hand Refinement stages, which PhysTwin does not perform.}
\label{tab:timecost}
\vspace{-0.5\baselineskip}
\end{table}

%% file: example_paper.bib
@article{romero2017embodied,
  title={Embodied hands: modeling and capturing hands and bodies together},
  author={Romero, Javier and Tzionas, Dimitrios and Black, Michael J},
  journal={ACM TOG},
  year={2017},
}

@article{liu2013fast,
  title={Fast simulation of mass-spring systems},
  author={Liu, Tiantian and Bargteil, Adam W and O'Brien, James F and Kavan, Ladislav},
  journal={ACM TOG},
  year={2013},
}

@inproceedings{zhang2020mediapipe,
title	= {MediaPipe: A framework for perceiving and processing reality},
author	= {Camillo Lugaresi and Jiuqiang Tang and Hadon Nash and Chris McClanahan and Esha Uboweja and Michael Hays and Fan Zhang and Chuo-Ling Chang and Ming Yong and Juhyun Lee and Wan-Teh Chang and Wei Hua and Manfred Georg and Matthias Grundmann},
year	= {2019},
booktitle	= {CVPRW}
}

@inproceedings{dong2024hamba,
  title={Hamba: Single-view 3d hand reconstruction with graph-guided bi-scanning mamba},
  author={Dong, Haoye and Chharia, Aviral and Gou, Wenbo and Vicente Carrasco, Francisco and De la Torre, Fernando D},
  booktitle={NeurIPS},
  year={2024}
}

@inproceedings{jiang2025phystwin,
  title={Phystwin: Physics-informed reconstruction and simulation of deformable objects from videos},
  author={Jiang, Hanxiao and Hsu, Hao-Yu and Zhang, Kaifeng and Yu, Hsin-Ni and Wang, Shenlong and Li, Yunzhu},
  booktitle={ICCV},
  year={2025}
}

@inproceedings{hasson2019learning,
  title={Learning joint reconstruction of hands and manipulated objects},
  author={Hasson, Yana and Varol, Gul and Tzionas, Dimitrios and Kalevatykh, Igor and Black, Michael J and Laptev, Ivan and Schmid, Cordelia},
  booktitle={CVPR},
  year={2019}
}

@inproceedings{mescheder2019occupancy,
  title={Occupancy networks: Learning 3d reconstruction in function space},
  author={Mescheder, Lars and Oechsle, Michael and Niemeyer, Michael and Nowozin, Sebastian and Geiger, Andreas},
  booktitle={CVPR},
  year={2019}
}

@inproceedings{chen2023gsdf,
  title={gSDF: Geometry-Driven Signed Distance Functions for 3D Hand-Object Reconstruction},
  author={Chen, Zerui and Chen, Shizhe and Schmid, Cordelia and Laptev, Ivan},
  booktitle={CVPR},
  year={2023}
}

@inproceedings{chao2021dexycb,
  title={DexYCB: A benchmark for capturing hand grasping of objects},
  author={Chao, Yu-Wei and Yang, Wei and Xiang, Yu and Molchanov, Pavlo and Handa, Ankur and Tremblay, Jonathan and Narang, Yashraj S and Van Wyk, Karl and Iqbal, Umar and Birchfield, Stan and others},
  booktitle={CVPR},
  year={2021}
}

@inproceedings{loshchilov2017decoupled,
  title={Decoupled weight decay regularization},
  author={Loshchilov, Ilya and Hutter, Frank},
  booktitle={CoRR},
  volume = {abs/1711.05101},
  year={2017}
}

@book{lozano2006towards,
  title={Towards a new evolutionary computation: advances on estimation of distribution algorithms},
  author={Lozano, Jose A},
  volume={192},
  year={2006}
}

@inproceedings{fan2023arctic,
    title = {{ARCTIC}: A Dataset for Dexterous Bimanual Hand-Object Manipulation},
    author = {Fan, Zicong and Taheri, Omid and Tzionas, Dimitrios and Kocabas, Muhammed and Kaufmann, Manuel and Black, Michael J. and Hilliges, Otmar},
    booktitle = {CVPR},
    year = {2023}
}

@inproceedings{hampali2020honnotate,
  title={Honnotate: A method for 3d annotation of hand and object poses},
  author={Hampali, Shreyas and Rad, Mahdi and Oberweger, Markus and Lepetit, Vincent},
  booktitle={CVPR},
  year={2020}
}

@inproceedings{kingma2014adam,
  title={Adam: A method for stochastic optimization},
  author={Kingma, Diederik P},
  booktitle={ICLR},
  year={2014}
}

@inproceedings{jung2025learning,
  title={Learning Dense Hand Contact Estimation from Imbalanced Data},
  author={Jung, Daniel Sungho and Lee, Kyoung Mu},
  booktitle={CoRR},
  volume={arXiv:2505.11152},
  year={2025}
}

@inproceedings{mueller2017real,
  title={Real-time hand tracking under occlusion from an egocentric rgb-d sensor},
  author={Mueller, Franziska and Mehta, Dushyant and Sotnychenko, Oleksandr and Sridhar, Srinath and Casas, Dan and Theobalt, Christian},
  booktitle={ICCV},
  year={2017}
}

@inproceedings{garcia2018first,
  title={First-person hand action benchmark with rgb-d videos and 3d hand pose annotations},
  author={Garcia-Hernando, Guillermo and Yuan, Shanxin and Baek, Seungryul and Kim, Tae-Kyun},
  booktitle={CVPR},
  year={2018}
}

@inproceedings{brahmbhatt2020contactpose,
  title={ContactPose: A dataset of grasps with object contact and hand pose},
  author={Brahmbhatt, Samarth and Tang, Chengcheng and Twigg, Christopher D and Kemp, Charles C and Hays, James},
  booktitle={ECCV},
  year={2020}
}

@inproceedings{taheri2020grab,
  title={GRAB: A dataset of whole-body human grasping of objects},
  author={Taheri, Omid and Ghorbani, Nima and Black, Michael J and Tzionas, Dimitrios},
  booktitle={ECCV 2020},
  year={2020}
}

@inproceedings{swamy2023showme,
  title={Showme: Benchmarking object-agnostic hand-object 3d reconstruction},
  author={Swamy, Anilkumar and Leroy, Vincent and Weinzaepfel, Philippe and Baradel, Fabien and Galaaoui, Salma and Br{\'e}gier, Romain and Armando, Matthieu and Franco, Jean-Sebastien and Rogez, Gr{\'e}gory},
  booktitle={ICCV},
  year={2023}
}

@inproceedings{corona2020ganhand,
  title={Ganhand: Predicting human grasp affordances in multi-object scenes},
  author={Corona, Enric and Pumarola, Albert and Alenya, Guillem and Moreno-Noguer, Francesc and Rogez, Gr{\'e}gory},
  booktitle={CVPR},
  year={2020}
}

@article{damen2022rescaling,
  title={Rescaling egocentric vision: Collection, pipeline and challenges for epic-kitchens-100},
  author={Damen, Dima and Doughty, Hazel and Farinella, Giovanni Maria and Furnari, Antonino and Kazakos, Evangelos and Ma, Jian and Moltisanti, Davide and Munro, Jonathan and Perrett, Toby and Price, Will and others},
  journal={IJCV},
  year={2022}
}

@inproceedings{brahmbhatt2019contactdb,
  title={Contactdb: Analyzing and predicting grasp contact via thermal imaging},
  author={Brahmbhatt, Samarth and Ham, Cusuh and Kemp, Charles C and Hays, James},
  booktitle={CVPR},
  year={2019}
}

@inproceedings{tse2022s,
  title={S2 Contact: Graph-Based Network for 3D Hand-Object Contact Estimation with Semi-supervised Learning},
  author={Tse, Tze Ho Elden and Zhang, Zhongqun and Kim, Kwang In and Leonardis, Ales and Zheng, Feng and Chang, Hyung Jin},
  booktitle={ECCV},
  year={2022}
}

@inproceedings{grady2021contactopt,
  title={Contactopt: Optimizing contact to improve grasps},
  author={Grady, Patrick and Tang, Chengcheng and Twigg, Christopher D and Vo, Minh and Brahmbhatt, Samarth and Kemp, Charles C},
  booktitle={CVPR},
  year={2021}
}

@inproceedings{liu2023contactgen,
  title={ContactGen: Generative Contact Modeling for Grasp Generation},
  author={Liu, Shaowei and Zhou, Yang and Yang, Jimei and Gupta, Saurabh and Wang, Shenlong},
  booktitle={ICCV},
  year={2023}
}

@article{chen2021joint,
  title={Joint hand-object 3d reconstruction from a single image with cross-branch feature fusion},
  author={Chen, Yujin and Tu, Zhigang and Kang, Di and Chen, Ruizhi and Bao, Linchao and Zhang, Zhengyou and Yuan, Junsong},
  journal={TIP},
  year={2021}
}

@inproceedings{liu2021semi,
  title={Semi-supervised 3d hand-object poses estimation with interactions in time},
  author={Liu, Shaowei and Jiang, Hanwen and Xu, Jiarui and Liu, Sifei and Wang, Xiaolong},
  booktitle={CVPR},
  year={2021}
}

@inproceedings{doosti2020hope,
  title={Hope-net: A graph-based model for hand-object pose estimation},
  author={Doosti, Bardia and Naha, Shujon and Mirbagheri, Majid and Crandall, David J},
  booktitle={CVPR},
  year={2020}
}

@inproceedings{hasson2020leveraging,
  title={Leveraging photometric consistency over time for sparsely supervised hand-object reconstruction},
  author={Hasson, Yana and Tekin, Bugra and Bogo, Federica and Laptev, Ivan and Pollefeys, Marc and Schmid, Cordelia},
  booktitle={CVPR},
  year={2020}
}

@inproceedings{hampali2022keypoint,
  title={Keypoint transformer: Solving joint identification in challenging hands and object interactions for accurate 3d pose estimation},
  author={Hampali, Shreyas and Sarkar, Sayan Deb and Rad, Mahdi and Lepetit, Vincent},
  booktitle={CVPR},
  year={2022}
}

@inproceedings{tekin2019h,
  title={H+o: Unified egocentric recognition of 3d hand-object poses and interactions},
  author={Tekin, Bugra and Bogo, Federica and Pollefeys, Marc},
  booktitle={CVPR},
  year={2019}
}

@inproceedings{chen2022alignsdf,
  title={Alignsdf: Pose-aligned signed distance fields for hand-object reconstruction},
  author={Chen, Zerui and Hasson, Yana and Schmid, Cordelia and Laptev, Ivan},
  booktitle={ECCV},
  year={2022}
}

@inproceedings{park20203d,
  title={3d hand pose estimation with a single infrared camera via domain transfer learning},
  author={Park, Gabyong and Kim, Tae-Kyun and Woo, Woontack},
  booktitle={ISMAR},
  year={2020}
}

@article{xie2023hmdo,
  title={Hmdo: Markerless multi-view hand manipulation capture with deformable objects},
  author={Xie, Wei and Yu, Zhipeng and Zhao, Zimeng and Zuo, Binghui and Wang, Yangang},
  journal={Graph. Models},
  year={2023}
}

@inproceedings{xie2023nonrigid,
  title={Nonrigid Object Contact Estimation With Regional Unwrapping Transformer},
  author={Wei Xie and Zimeng Zhao and Shiying Li and Binghui Zuo and Yangang Wang},
  booktitle={ICCV},
  year={2023},
}

@article{qi2025human,
  title={Human Grasp Generation for Rigid and Deformable Objects with Decomposed VQ-VAE},
  author={Qi, Mengshi and Zhao, Zhe and Ma, Huadong},
  journal={arXiv preprint arXiv:2501.05483},
  year={2025}
}

@inproceedings{zhu2024contactart,
  title={Contactart: Learning 3d interaction priors for category-level articulated object and hand poses estimation},
  author={Zhu, Zehao and Wang, Jiashun and Qin, Yuzhe and Sun, Deqing and Jampani, Varun and Wang, Xiaolong},
  booktitle={3DV},
  year={2024}
}

@inproceedings{zhang2025bimart,
  title={Bimart: A unified approach for the synthesis of 3d bimanual interaction with articulated objects},
  author={Zhang, Wanyue and Dabral, Rishabh and Golyanik, Vladislav and Choutas, Vasileios and Alvarado, Eduardo and Beeler, Thabo and Habermann, Marc and Theobalt, Christian},
  booktitle={CVPR},
  year={2025}
}

@inproceedings{mildenhall2020nerf,
  title={NeRF: Representing Scenes as Neural Radiance Fields for View Synthesis},
  author={Mildenhall, Ben and Srinivasan, Pratul P and Tancik, Matthew and Barron, Jonathan T and Ramamoorthi, Ravi and Ng, Ren},
  booktitle={ECCV},
  year={2020}
}

@inproceedings{attal2023hyperreel,
  title={Hyperreel: High-fidelity 6-dof video with ray-conditioned sampling},
  author={Attal, Benjamin and Huang, Jia-Bin and Richardt, Christian and Zollhoefer, Michael and Kopf, Johannes and O’Toole, Matthew and Kim, Changil},
  booktitle={CVPR},
  year={2023}
}

@inproceedings{kratimenos2024dynmf,
  title={Dynmf: Neural motion factorization for real-time dynamic view synthesis with 3d gaussian splatting},
  author={Kratimenos, Agelos and Lei, Jiahui and Daniilidis, Kostas},
  booktitle={ECCV},
  year={2024}
}

@inproceedings{li2023dynibar,
  title={Dynibar: Neural dynamic image-based rendering},
  author={Li, Zhengqi and Wang, Qianqian and Cole, Forrester and Tucker, Richard and Snavely, Noah},
  booktitle={CVPR},
  year={2023}
}

@inproceedings{luiten2024dynamic,
  title={Dynamic 3d gaussians: Tracking by persistent dynamic view synthesis},
  author={Luiten, Jonathon and Kopanas, Georgios and Leibe, Bastian and Ramanan, Deva},
  booktitle={3DV},
  year={2024}
}

@inproceedings{park2021nerfies,
  title={Nerfies: Deformable neural radiance fields},
  author={Park, Keunhong and Sinha, Utkarsh and Barron, Jonathan T and Bouaziz, Sofien and Goldman, Dan B and Seitz, Steven M and Martin-Brualla, Ricardo},
  booktitle={ICCV},
  year={2021}
}

@article{park2021hypernerf,
  title={HyperNeRF: a higher-dimensional representation for topologically varying neural radiance fields},
  author={Park, Keunhong and Sinha, Utkarsh and Hedman, Peter and Barron, Jonathan T and Bouaziz, Sofien and Goldman, Dan B and Martin-Brualla, Ricardo and Seitz, Steven M},
  journal={ACM TOG},
  year={2021}
}

@inproceedings{pumarola2021d,
  title={D-nerf: Neural radiance fields for dynamic scenes},
  author={Pumarola, Albert and Corona, Enric and Pons-Moll, Gerard and Moreno-Noguer, Francesc},
  booktitle={CVPR},
  year={2021}
}

@inproceedings{wang2023flow,
  title={Flow supervision for deformable nerf},
  author={Wang, Chaoyang and MacDonald, Lachlan Ewen and Jeni, Laszlo A and Lucey, Simon},
  booktitle={CVPR},
  year={2023}
}

@inproceedings{xian2021space,
  title={Space-time neural irradiance fields for free-viewpoint video},
  author={Xian, Wenqi and Huang, Jia-Bin and Kopf, Johannes and Kim, Changil},
  booktitle={CVPR},
  year={2021}
}

@inproceedings{yu2023dylin,
  title={Dylin: Making light field networks dynamic},
  author={Yu, Heng and Julin, Joel and Milacski, Zoltan A and Niinuma, Koichiro and Jeni, L{\'a}szl{\'o} A},
  booktitle={CVPR},
  year={2023}
}

@inproceedings{tretschk2021nonrigid,
  title={Non-rigid neural radiance fields: Reconstruction and novel view synthesis of a dynamic scene from monocular video},
  author={Tretschk, Edgar and Tewari, Ayush and Golyanik, Vladislav and Zollh{\"o}fer, Michael and Lassner, Christoph and Theobalt, Christian},
  booktitle={ICCV},
  year={2021}
}

@article{chu2022physics,
  title={Physics informed neural fields for smoke reconstruction with sparse data},
  author={Chu, Mengyu and Liu, Lingjie and Zheng, Quan and Franz, Aleksandra and Seidel, Hans-Peter and Theobalt, Christian and Zayer, Rhaleb},
  journal={ACM TOG},
  year={2022}
}

@inproceedings{curless1996volumetric,
  title={A volumetric method for building complex models from range images},
  author={Curless, Brian and Levoy, Marc},
  booktitle={SIGGRAPH},
  year={1996}
}

@inproceedings{li2008global,
  title={Global correspondence optimization for non-rigid registration of depth scans},
  author={Li, Hao and Sumner, Robert W and Pauly, Mark},
  booktitle={CGF},
  year={2008},
}

@inproceedings{newcombe2015dynamicfusion,
  title={Dynamicfusion: Reconstruction and tracking of non-rigid scenes in real-time},
  author={Newcombe, Richard A and Fox, Dieter and Seitz, Steven M},
  booktitle={CVPR},
  year={2015}
}

@article{wang2015deformation,
  title={Deformation capture and modeling of soft objects.},
  author={Wang, Bin and Wu, Longhua and Yin, KangKang and Ascher, Uri M and Liu, Libin and Huang, Hui},
  journal={ACM TOG},
  year={2015}
}

@inproceedings{qiao2021differentiable,
  title={Differentiable simulation of soft multi-body systems},
  author={Qiao, Yiling and Liang, Junbang and Koltun, Vladlen and Lin, Ming},
  booktitle={NeurIPS},
  year={2021}
}

@article{du2021diffpd,
  title={Diffpd: Differentiable projective dynamics},
  author={Du, Tao and Wu, Kui and Ma, Pingchuan and Wah, Sebastien and Spielberg, Andrew and Rus, Daniela and Matusik, Wojciech},
  journal={ACM TOG},
  year={2021}
}

@article{geilinger2020add,
  title={Add: Analytically differentiable dynamics for multi-body systems with frictional contact},
  author={Geilinger, Moritz and Hahn, David and Zehnder, Jonas and B{\"a}cher, Moritz and Thomaszewski, Bernhard and Coros, Stelian},
  journal={ACM TOG},
  year={2020}
}

@inproceedings{jatavallabhula2021gradsim,
  title={gradsim: Differentiable simulation for system identification and visuomotor control},
  author={Murthy, J Krishna and Macklin, Miles and Golemo, Florian and Voleti, Vikram and Petrini, Linda and Weiss, Martin and Considine, Breandan and Parent-L{\'e}vesque, J{\'e}r{\^o}me and Xie, Kevin and Erleben, Kenny and others},
  booktitle={ICLR},
  year={2020}
}

@inproceedings{zhang2024physdreamer,
  title={Physdreamer: Physics-based interaction with 3d objects via video generation},
  author={Zhang, Tianyuan and Yu, Hong-Xing and Wu, Rundi and Feng, Brandon Y and Zheng, Changxi and Snavely, Noah and Wu, Jiajun and Freeman, William T},
  booktitle={ECCV},
  year={2024}
}

@inproceedings{li2023pac,
  title={PAC-NeRF: Physics Augmented Continuum Neural Radiance Fields for Geometry-Agnostic System Identification},
  author={Li, Xuan and Qiao, Yi-Ling and Chen, Peter Yichen and Jatavallabhula, Krishna Murthy and Lin, Ming and Jiang, Chenfanfu and Gan, Chuang},
  booktitle={ICLR},
  year={2023}
}

@inproceedings{chen2022virtual,
  title={Virtual elastic objects},
  author={Chen, Hsiao-yu and Tretschk, Edith and Stuyck, Tuur and Kadlecek, Petr and Kavan, Ladislav and Vouga, Etienne and Lassner, Christoph},
  booktitle={CVPR},
  year={2022}
}

@inproceedings{zhong2024reconstruction,
  title={Reconstruction and simulation of elastic objects with spring-mass 3d gaussians},
  author={Zhong, Licheng and Yu, Hong-Xing and Wu, Jiajun and Li, Yunzhu},
  booktitle={ECCV},
  year={2024}
}

@inproceedings{qiao2022neuphysics,
  title={Neuphysics: Editable neural geometry and physics from monocular videos},
  author={Qiao, Yi-Ling and Gao, Alexander and Lin, Ming},
  booktitle={NeurIPS},
  year={2022}
}

@inproceedings{xiang2025structured,
  title={Structured 3d latents for scalable and versatile 3d generation},
  author={Xiang, Jianfeng and Lv, Zelong and Xu, Sicheng and Deng, Yu and Wang, Ruicheng and Zhang, Bowen and Chen, Dong and Tong, Xin and Yang, Jiaolong},
  booktitle={CVPR},
  year={2025}
}

@article{karaev2024cotracker3,
  title={Cotracker3: Simpler and better point tracking by pseudo-labelling real videos},
  author={Karaev, Nikita and Makarov, Iurii and Wang, Jianyuan and Neverova, Natalia and Vedaldi, Andrea and Rupprecht, Christian},
  booktitle={CoRR},
  volume = {arXiv:2410.11831},
  year={2024}
}

@inproceedings{li2021hybrik,
  title={Hybrik: A hybrid analytical-neural inverse kinematics solution for 3d human pose and shape estimation},
  author={Li, Jiefeng and Xu, Chao and Chen, Zhicun and Bian, Siyuan and Yang, Lixin and Lu, Cewu},
  booktitle={CVPR},
  year={2021}
}

@inproceedings{nealen2006physically,
  title={Physically based deformable models in computer graphics},
  author={Nealen, Andrew and M{\"u}ller, Matthias and Keiser, Richard and Boxerman, Eddy and Carlson, Mark},
  booktitle={Computer graphics forum},
  year={2006}
}

@article{silling2005meshfree,
  title={A meshfree method based on the peridynamic model of solid mechanics},
  author={Silling, Stewart A and Askari, Ebrahim},
  journal={Computers \& structures},
  year={2005}
}

@article{silling2007peridynamic,
  title={Peridynamic states and constitutive modeling},
  author={Silling, Stewart A and Epton, Michael and Weckner, Olaf and Xu, Jifeng and Askari, E23481501120},
  journal={Journal of elasticity},
  year={2007}
}

@article{wang2023determination,
  title={Determination of horizon size in state-based peridynamics},
  author={Wang, Bingquan and Oterkus, Selda and Oterkus, Erkan},
  journal={Continuum Mechanics and Thermodynamics},
  year={2023}
}

@inproceedings{zhang2024dynamics,
  title={Dynamic 3D Gaussian Tracking for Graph-Based Neural Dynamics Modeling},
  author={Zhang, Mingtong and Zhang, Kaifeng and Li, Yunzhu},
  booktitle={CoRL},
  year={2024}
}

@inproceedings{xu2026neuspring,
  title={NeuSpring: Neural Spring Fields for Reconstruction and Simulation of Deformable Objects from Videos},
  author={Qingshan Xu and Jiao Liu and Shangshu Yu and Yuxuan Wang and Yuan Zhou and Junbao Zhou and Jiequan Cui and Yew-Soon Ong and Hanwang Zhang},
  booktitle={AAAI},
  year={2026}
}

@article{yang2025physworld,
  title={PhysWorld: From Real Videos to World Models of Deformable Objects via Physics-Aware Demonstration Synthesis},
  author={Yu Yang and Zhilu Zhang and Xiang Zhang and Yihan Zeng and Hui Li and Wangmeng Zuo},
  journal={arXiv preprint arXiv:2510.21447},
  year={2025}
}

@inproceedings{cho2024dense,
  title={Dense hand-object (ho) graspnet with full grasping taxonomy and dynamics},
  author={Cho, Woojin and Lee, Jihyun and Yi, Minjae and Kim, Minje and Woo, Taeyun and Kim, Donghwan and Ha, Taewook and Lee, Hyokeun and Ryu, Je-Hwan and Woo, Woontack and others},
  booktitle={ECCV},
  year={2024}
}

@inproceedings{lee2024interhandgen,
  title={InterHandGen: Two-Hand Interaction Generation via Cascaded Reverse Diffusion},
  author={Lee, Jihyun and Saito, Shunsuke and Nam, Giljoo and Sung, Minhyuk and Kim, Tae-Kyun},
  booktitle={CVPR},
  year={2024}
}

@inproceedings{garcia2020physics,
  title={Physics-based dexterous manipulations with estimated hand poses and residual reinforcement learning},
  author={Garcia-Hernando, Guillermo and Johns, Edward and Kim, Tae-Kyun},
  booktitle={IROS},
  year={2020}
}

@inproceedings{antotsiou2021adversarial,
  title={Adversarial Imitation Learning with Trajectorial Augmentation and Correction},
  author={Dafni Antotsiou and Carlo Ciliberto and Tae-Kyun Kim},
  year={2021},
  booktitle={ICRA},
}

@inproceedings{kim2024bitt,
  title={BiTT: Bi-directional Texture Reconstruction of Interacting Two Hands from a Single Image},
  author={Kim, Minje and Kim, Tae-Kyun},
  booktitle={CVPR},
  year={2024}
}

@inproceedings{lee2023im2hands,
  title={Im2hands: Learning attentive implicit representation of interacting two-hand shapes},
  author={Lee, Jihyun and Sung, Minhyuk and Choi, Honggyu and Kim, Tae-Kyun},
  booktitle={CVPR},
  year={2023}
}

@inproceedings{kim2024mhcdiff,
  title={Multi-hypotheses Conditioned Point Cloud Diffusion for 3D Human Reconstruction from Occluded Images},
  author={Donghwan Kim and Tae-Kyun Kim},
  booktitle={NeurIPS},
  year={2024}
}
